\documentclass[a4paper, 10pt, conference]{ieeeconf}      

\IEEEoverridecommandlockouts                              
\usepackage[utf8]{inputenc}
\usepackage[T1]{fontenc}
\usepackage{cite}
\usepackage[pdftex]{graphicx}
\usepackage{graphicx}
\usepackage{color} 
\usepackage{everyshi} 
\usepackage{subfig} 
\usepackage{booktabs} 
\usepackage{multirow} 
\usepackage{threeparttable} 
\usepackage{bm} 
\usepackage{amsmath}
\usepackage{amssymb} 
\usepackage{amsfonts} 
\usepackage{pifont} 
   
\title{\LARGE \bf
Demonstrations of Cooperative Perception: Safety and Robustness in Connected and Automated Vehicle Operations
}

\author{Mao Shan, Karan Narula, Yung Fei Wong, Stewart Worrall, Malik Khan, Paul Alexander and Eduardo Nebot
\thanks{M. Shan, K. Narula, S. Worrall and E. Nebot are with the Australian Centre for Field Robotics, The University of Sydney, NSW 2006, Australia. E-mails: {\tt\small \{mao.shan, karan.narula, stewart.worrall, eduardo.nebot\}@sydney.edu.au}}
\thanks{Y.F. Wong, M. Khan and P. Alexander are with Cohda Wireless, 27 Greenhill Road, Wayville, SA 5034, Australia. E-mails: {\tt\small \{ricky.wong, malik.khan, paul.alexander\}@cohdawireless.com}}
\thanks{This research is funded by iMOVE CRC and supported by the Cooperative Research Centres program, an Australian Government initiative.}
}

\begin{document}

\maketitle
\thispagestyle{empty}
\pagestyle{empty}

\begin{abstract}
Cooperative perception, or collective perception (CP) is an emerging and promising technology for intelligent transportation systems (ITS). It enables an ITS station (ITS-S) to share its local perception information with others by means of vehicle-to-X (V2X) communication, thereby achieving improved efficiency and safety in road transportation. In this paper, we present our recent progress on the development of a connected and automated vehicle (CAV) and intelligent roadside unit (IRSU). We present three different experiments to demonstrate the use of CP service within intelligent infrastructure to improve awareness of vulnerable road users (VRU) and thus safety for CAVs in various traffic scenarios. We demonstrate in the experiments that a connected vehicle (CV) can ``see'' a pedestrian around the corners. More importantly, we demonstrate how CAVs can autonomously and safely interact with walking and running pedestrians, relying only on the CP information from the IRSU through vehicle-to-infrastructure (V2I) communication. This is one of the first demonstrations of urban vehicle automation using only CP information. We also address in the paper the handling of collective perception messages (CPMs) received from the IRSU, and passing them through a pipeline of CP information coordinate transformation with uncertainty, multiple road user tracking, and eventually path planning/decision making within the CAV. The experimental results were obtained with manually driven CV, fully autonomous CAV, and an IRSU retrofitted with vision and laser sensors and a road user tracking system.
\end{abstract}

\section{Introduction}
\label{sec:introduction}

Autonomous vehicles (AVs) have received extensive attention in recent years as a rapidly emerging and disruptive technology to improve safety and efficiency of current road transportation systems. Most of the existing and under development AVs rely on local sensors, such as cameras and lidars, to perceive the environment and interact with other road users. Despite significant advances in sensor technology in recent years, the perception capability of these local sensors is ultimately bounded in range and field of view (FOV) due to their physical constraints. Besides, occluding objects in urban traffic environments such as buildings, trees, and other road users impose challenges in perception. There are also robustness related concerns, for instance, sensor degradation in adverse weather conditions, sensor interference, hardware malfunction and failure. Unfortunately, failing to maintain sufficient awareness of other road users, vulnerable road users (VRU) in particular, can cause catastrophic safety consequences for AVs.

In recent years, V2X communication has garnered increasing popularity among researchers in the field of intelligent transportation system (ITS) and with automobile manufacturers, as it enables a vehicle to share essential information with other road users in a V2X network. This can be a game changer for both human operated and autonomous vehicles, which would be referred to as connected vehicles (CVs) and connected and automated vehicles (CAVs), respectively. It will also open many doors to new possibilities with peer-to-peer connectivity. The connected agents within the cooperative ITS (C-ITS) network will be able to exploit the significant benefits that come from sharing information amongst the network. For instance, the standardised cooperative awareness messages (CAMs) enable mutual awareness between connected agents. Nevertheless, there are other types of road users such as non-connected vehicles, pedestrians, and cyclists that have not been included in the C-ITS services yet. The detection of these non-connected road users in this case becomes an important task for road safety.

The major standardisation organisations such as European Telecommunications Standard Institute (ETSI), SAE and IEEE have made a significant effort to standardise specifications regarding C-ITS services, V2X communication protocols, and security. This is essential to facilitate the deployment of C-ITS in road transportation network globally. The collective perception (CP) service is among those C-ITS services that are currently being standardised by ETSI. The CP service enables an ITS station (ITS-S), for instance, a CAV or an intelligent roadside unit (IRSU) to share its perception information with adjacent ITS-Ss by exchanging Collective Perception Messages (CPMs) via V2X communication. The ETSI CPMs convey abstract representations of perceived objects instead of raw sensory data, facilitating the interoperability between ITS-Ss of different types and from different manufactures. A CAV can benefit from the CP service in terms of improved awareness of surrounding road users, which is essential for ensuring road safety. Specifically, it facilitates a CAV to extend its sensing range and improve sensing quality, redundancy, and robustness through cross-platform sensor fusion, i.e., fusing its local sensory data with other CAVs and IRSUs information. Besides, the improved perception quality as a result of the sensor fusion potentially relaxes the accuracy and reliability requirements of onboard sensors. This could lower per vehicle cost to facilitate the massive deployment of CAV technology. As for traditional vehicles, CP also brings an attractive advantage of enabling perception capability without retrofitting the vehicle with perception sensors and the associated processing unit.

Over the last two years, we have been working with Cohda Wireless on CP and CAV. We are particularly interested in the safety implications the CP service is bringing into the current and future transportation network, and how the CP service will potentially shape the development of intelligent vehicles. To this end, we have developed an IRSU and tested it with the Australian Centre for Field Robotics (ACFR) CAV platforms in different traffic scenarios. This paper includes three representative experiments to showcase how a CV and a CAV achieves improved safety and robustness when perceiving and interacting with VRU using the CP information from an intelligent infrastructure in different traffic environments and with different setups.

The vehicles used are equipped with a suite of local perception sensors to implement full autonomy. Nevertheless, the current experiments do not use the internal perception capabilities so as to highlight the benefits of using intelligent infrastructure and CP service in the traffic environments. The received perception data from IRSU is used as the only or main source of information for multiple road user tracking and path planning within the smart vehicle we tested with. The first experiment was conducted on a public road in an urban traffic environment and the CV was able to ``see'' a visually obstructed pedestrian before making a turn to an alleyway. It is demonstrated in the next two experiments a CAV navigated autonomously and safely when interacting with walking and running pedestrians in proximity in a simulated and real lab traffic environments, respectively. Lastly, the paper also addresses the coordinate transformation of perception information considering the respective uncertainties, and analyses the influence of ITS-S self-localisation accuracy through numerical simulations.

The remainder of the paper is organised as follows. Section \ref{sec:related_work} will focus on the related work on CP and its use cases for CAV. Section \ref{sec:platforms} presents the IRSU and CAV platforms developed and used in the experiments, and the cross-platform perception information transformation with uncertainty is addressed in Section \ref{sec:transformation}. The results from the three experiments are presented in Section \ref{sec:results}, followed by conclusions drawn in Section \ref{sec:conclusions}. 

\section{Related Work}
\label{sec:related_work}

The concept of CP has been extensively studied in the ITS research community over the last two decades. Initial CP related work proposes to share raw sensory data between two mobile agents, such as images \cite{paper:LiNashashibi2011}, lidar point clouds \cite{paper:KumarShi2012}, both combined \cite{paper:MourllionLambert2004, paper:KimChong2013, paper:KimQin2015}, location and relative range measurements \cite{paper:KaramChausse2006, paper:ShanWorrall2014}. Those approaches however tend to require prohibitively high bandwidth for existing V2X communication technologies in a dense environment. Besides, raw sensor data is often vendor dependent and proprietary, causing interoperability issues among communicating ITS-Ss.

More theoretical and experimental work on CP was conducted as part of the Ko-FAS \cite{web:KoFAS} research initiative. These include \cite{paper:RauchKlanner2011, paper:RauchKlanner2012, paper:RauchMaier2013} based on the Ko-PER specified Cooperative Perception Message (CPM). It is a supplementary message to the standard ETSI ITS G5 CAMs, to support the abstract description of perceived dynamic and static objects. Experimental studies are conducted in \cite{paper:RauchKlanner2011} on the Ko-PER CPM transmission latency and range. In \cite{paper:RauchKlanner2012}, a high level object fusion framework in CP is proposed, which combines the local sensor information with the perception data received from other V2X enabled vehicles or roadside units (RSUs). Reference \cite{paper:RauchMaier2013} investigates the inter-vehicle data association and fusion for CP. More recent work in \cite{paper:AlligWanielik2019ITSC} proposes a variant of Ko-PER CPM and analyses the trade-off between message size as a result of enabling optional data fields in the CPM and global fusion accuracy.

Based on the work in \cite{paper:RauchKlanner2011}, reference \cite{paper:GuntherMennenga2016} proposes Environmental Perception Message (EPM) for CP with different information containers specifying sensor characteristics, originating station state, and parameters of perceived objects. It is also addressed in \cite{paper:GuntherMennenga2016} high level object fusion using the perceived information in received EPMs. Both EPM and the earlier Ko-PER CPM are a separate message that contains all CP related data elements (DEs) and data frames (DFs), and has to be transmitted in parallel with an ETSI CAM. There is also work towards extended CAM. For instance, Cooperative Sensing Message (CSM) from AutoNet2030 \cite{web:AutoNet2030, report:Hobert2015, paper:HobertFestag2015} extends CAM specifications to include description of objects perceived by local or remote sensors. Following a similar concept of CP, Proxy CAM is presented in \cite{paper:KitazatoTsukada2016, paper:Tsukada2017, paper:TsukadaOi2020}, where intelligent infrastructure generates standard CAMs for those perceived road users, while the work in \cite{thesis:Burgstahler2017} proposes a CPM comprised of a collection of CAMs, each describing a perceived object. The work in \cite{paper:GuntherRiebl2016} and \cite{paper:GuntherRiebl2018} evaluates different EPM dissemination variants under low and high traffic densities and proposes to attach the CP relevant DFs in EPM to CAM to minimise communication overhead. The CPM currently being specified at ETSI, as in \cite{report:ETSI_CPS}, is derived from optimising the EPM and combining it with CAM. It is therefore more self-contained, no longer dependent on the reception of CAMs.

Similarly, there are early stage standardisation activities in SAE advanced application technical committee to standardise messages and protocols for sensor data sharing in SAE J3224 \cite{report:SAE_J3224}. These messages and protocols are not yet defined and are thus not considered in this work.

Considering the limited communication bandwidth and avoiding congestion in the wireless channel, more recent studies in the CP area tend to focus on the communication aspect of the technology, weighing up provided CP service quality and the V2X network resources. The work in \cite{paper:GarlichsGunther2019} investigates ETSI CPM generation rules that balance provided service quality and the V2X channel load. Reference \cite{paper:ThandavarayanSepulcre2019} provides an in-depth study on the impact of different CPM generation rules from the perspectives of V2X communication performance and perception capabilities in low and high density traffic scenarios. The authors of \cite{paper:HuangFang2019} raise the concern of redundant data sharing in V2V based CP with the increase of CAV penetration rate. To tackle the redundant transmission issue, a probabilistic data selection approach is presented in \cite{paper:HuangLi2020}. Reference \cite{paper:ThandavarayanSepulcre2019arxiv} proposes an adaptive CPM generation rule considering change in perceived object's state, and the authors of \cite{paper:DeloozFestag2019} propose to employ object filtering schemes in CPM, to improve communication performance while minimising the detriment to perception quality. Similarly, the work in \cite{paper:AokiHiguchi2020} presents a deep reinforcement learning based approach that a vehicle can employ when selecting data to transmit in CP to alleviate communication congestion.

There is also work conducted to explore the use cases, benefits, and challenges of CP. Reference \cite{paper:GuntherTrauer2015} provides early study of CP illustrating its potential in terms of improved vehicle awareness and extended perception range and field of view. The work presented in \cite{paper:GuntherMennenga2016} evaluates EPM for obstacle avoidance of two manually driven CAVs, showing that the CP helps gain extra reaction time for the vehicles to avoid obstacles. Reference \cite{paper:HuangFang2019} analyses the performance gain in extending horizon of CAVs by leveraging V2V based CP. The work in \cite{paper:SchieggBrahmi2019} and \cite{paper:SchieggBischoff2020} analytically evaluates the enhancement of environmental perception for CVs at different CP service penetration rates and with different traffic densities. The authors of \cite{paper:BianZhang2018} discuss the security threats in CP and propose possible countermeasures in V2X network protocols, while the work in \cite{paper:AmbrosinYang2019} focuses on using CP for detecting vehicle misbehaviour due to adversarial attacks in V2X communication. Most of CP related use cases studied in the literature are safety related, including cooperative driving \cite{paper:KimQin2015, paper:HobertFestag2015}, cooperative advisory warnings \cite{paper:SeeligerWeidl2014, paper:NaujoksGrattenthaler2015}, cooperative collision avoidance \cite{paper:KimChong2013, paper:GuntherMennenga2016, paper:DengDi2018}, intersection assistance \cite{paper:KitazatoTsukada2016, paper:RondinoneWalter2018}, and vehicle misbehaviour detection \cite{paper:KamelAnsari2020}, to name a few. It is presented in \cite{paper:WangVeciana2018} quantitative comparison of V2V and V2I connectivity on improving sensing redundancy and collaborative sensing coverage for CAV applications. The work concludes that infrastructure support is crucial for safety related services such as CP, especially when the penetration rate of sensing vehicles is low. The authors of \cite{paper:KitazatoTsukada2016} demonstrate improved awareness of approaching vehicles at an intersection using the CP information from an IRSU. The CP in the work is achieved through Proxy CAM. Reference \cite{paper:MerdrignacShagdar2018} compares CAM and CPM and demonstrates IRSU assisted augmented perception through simulations. Recent work in \cite{paper:JandialMerdrignac2019} demonstrates the IRSU assisted CP for extending perception of CAVs on open roads. Infrastructure-assisted CP is also part of the scope of Managing Automated Vehicles Enhances Network (MAVEN) \cite{web:MAVEN}, an EU funded project targeting traffic management solutions where CAVs are guided at signalised cooperative intersections in urban traffic environments \cite{paper:RondinoneWalter2018}. Other CP related joint research projects include TransAID \cite{paper:KhanAndert2019} and IMAGinE \cite{paper:LlasterMichalke2019}.

A significant proportion of the existing work conducts the analyses of V2X communication and CP in simulated environments. For instance, the work in \cite{paper:HuangFang2019, paper:HuangLi2020} is carried out in an open source microscopic road traffic simulation package SUMO (Simulation of Urban Mobility) \cite{web:SUMO}. Another commonly used network and mobility simulator is \textit{Veins} (Vehicles in Network Simulation) \cite{web:VEINS, paper:SommerGerman2011}, which integrates SUMO and a discrete-event simulator OMNeT++ (Open Modular Network Testbed in C++) \cite{web:OMNet} for modelling realistic communication patterns. The authors of \cite{paper:GuntherTrauer2015, paper:GuntherRiebl2016, paper:GuntherRiebl2018, paper:GarlichsWegner2018, paper:GarlichsGunther2019, paper:DeloozFestag2019} conduct work in \textit{Artery} \cite{web:Artery, paper:RieblGunther2015} framework, which wraps SUMO and OMNet++, and enables V2X simulations based on ETSI ITS G5 protocol stack. There are also simulators for advanced driving assistance systems (ADASs) and autonomous driving systems, which provide more realistic sensory level perception information. In recent years, they start to show their potential in testing and validating CP with CAVs. For instance, Pro-SiVIC is employed in \cite{paper:MerdrignacShagdar2018} for CP related simulations, and CARLA \cite{web:CARLA, paper:DosovitskiyRos2017} is combined with SUMO in a simulation platform developed for CP in \cite{paper:AokiHiguchi2020}.

\section{Platforms}
\label{sec:platforms}

\subsection{Intelligent Roadside Unit}
\label{sec:irsu}

\begin{figure*}[!t]
	\centering
	\includegraphics[width=4.0in]{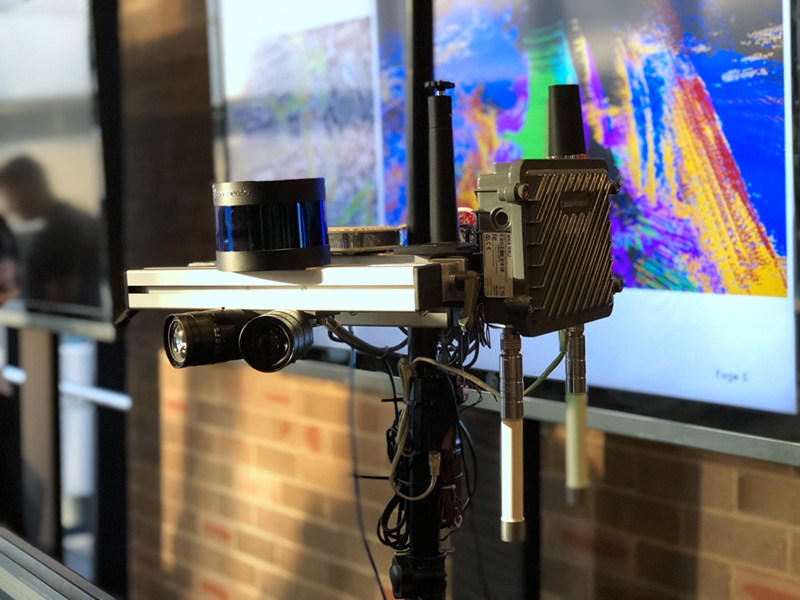}
	\caption{The developed IRSU is equipped with multiple sensors including dual cameras and a 16-beam lidar. The sensors sit on a tripod for easy deployment in the field.}
	\label{fig:irsu_sensors}
\end{figure*}

The IRSU developed comprises of a sensor head, a processing workstation, and a Cohda Wireless MK5 RSU. The sensor head is mounted on a tripod for easy deployment in different testing environments, as shown in Figure \ref{fig:irsu_sensors}. Specifically, the two Pointgrey Blackfly BFLY-PGE-23S6C-C cameras are mounted with an angle separation of $45^{\circ}$. The camera with its lens Fujinon CF12.5HA-1 has a horizontal field-of-view (FOV) of about $54^{\circ}$, and a vertical FOV of $42^{\circ}$. The setup of dual cameras achieves a combined FOV of approximately $100^{\circ}$, and the FOV can be further augmented by adding more cameras to the sensor head. A 16-beam lidar is also installed to the sensor head. The workstation has AMD Ryzen Threadripper 2950X CPU, 32GB memory, RTX2080Ti GPU, running Robot Operation System (ROS) Melodic on Ubuntu 18.04.2 long-term support (LTS).

\begin{figure*}[!t]
	\centering
	\includegraphics[width=5.0in]{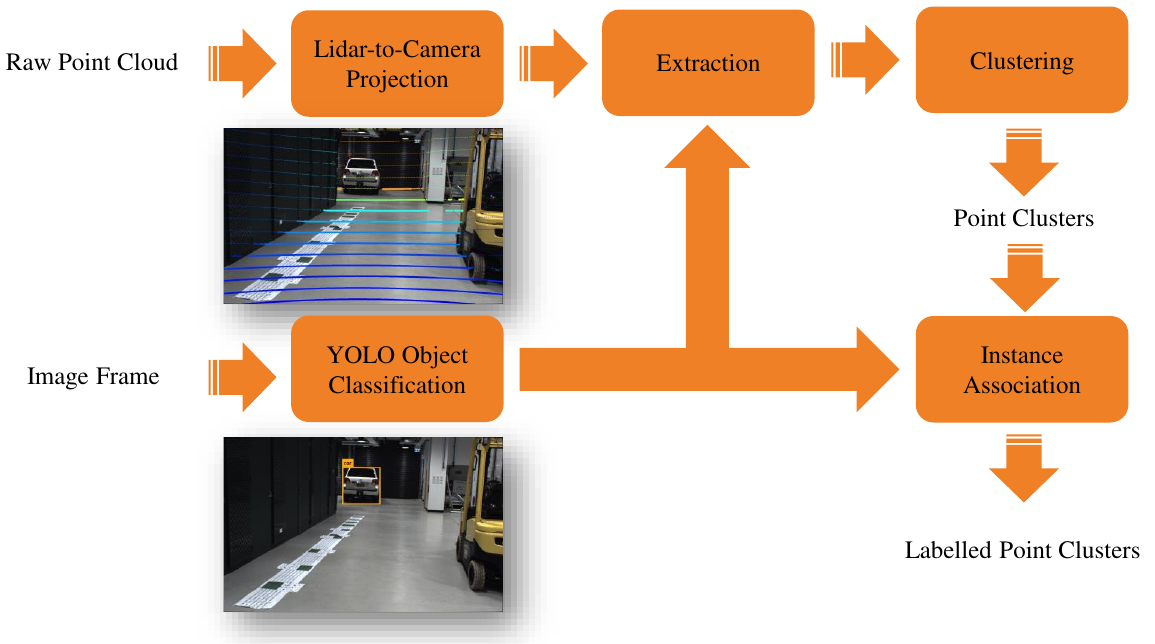}
	\caption{The sensory data processing pipeline within the IRSU.}
	\label{fig:irsu_perception_pipeline}
\end{figure*}

In terms of information processing, the workstation first processes the sensory data of images and lidar point clouds for pedestrian and vehicle detection. Specifically, the raw images from cameras are first rectified using camera intrinsic calibration parameters. As illustrated in Figure \ref{fig:irsu_perception_pipeline}, the road users within the images are classified/detected using YOLOv3 that runs on GPU. At the same time the lidar point clouds are projected to the image coordinate system with proper extrinsic sensor calibration parameters. The lidar points are then segmented, clustered and labelled by fusing the visual classifier results (in the form of bounding boxes in the image) and the projected lidar points.

The detection results are then encoded into ETSI CPMs and broadcast by Cohda MK5 at 10 Hz. Details are available in Section \ref{sec:codec}. We tested the working range of the developed IRSU for detecting common road users, such as pedestrians and vehicles. The maximum detection range is approximately 20 m for pedestrians and 40 m for cars. Also in the IRSU, a variant of Gaussian mixture probability hypothesis density (GMPHD) filter \cite{paper:VoMa2006} is employed to track multiple road users and has its tracking results visualised in real time within the workstation. The same tracking algorithm is also employed on the CAV side. Details would be given in Section \ref{sec:cav}.


\begin{figure*}[!t]
	\centering
	\subfloat[]{ 
		\label{fig:irsu_ped_tracking_setup:a} 
		\includegraphics[width=2.4in]{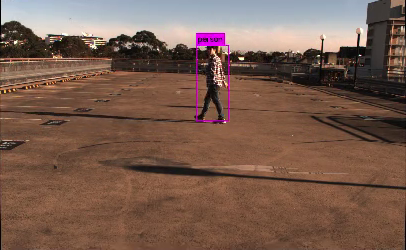}}
	\subfloat[]{ 
		\label{fig:irsu_ped_tracking_setup:b} 
		\includegraphics[width=2.4in]{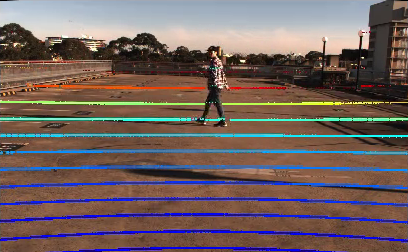}}
	\vfill
	\subfloat[]{ 
		\label{fig:irsu_ped_tracking_setup:c} 
		\includegraphics[width=4.9in]{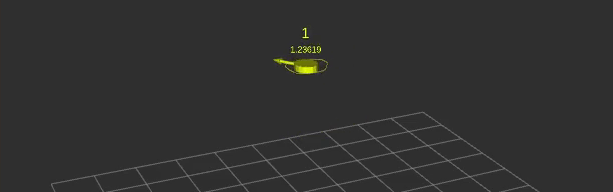}}
	\caption{Pedestrian tracking setup for the IRSU. (a) shows the detection of the target pedestrian within a camera image using YOLOv3. The lidar points are projected to the image frame with extrinsic calibration parameters. The projected points in (b) are colour coded based on their range to the sensor in 3D space. Those bold points indicate those hitting the ground plane. (c) demonstrates the tracking of the pedestrian in 3D space. The RTK antenna was hidden within the cap of the pedestrian to log GNSS positions as the ground truth.}
	\label{fig:irsu_ped_tracking_setup} 
\end{figure*}


\begin{figure*}[!t]
    \centering
    \begin{minipage}[c]{3.1in}
        \centering
        \subfloat[]{
            \label{fig:irsu_ped_tracking_results:a} 
            \includegraphics[width=3.0in]{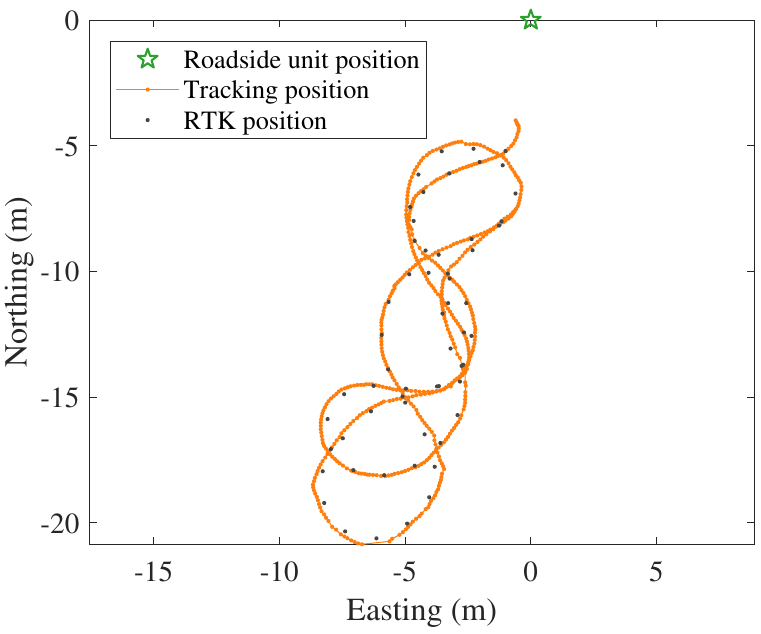}}
    \end{minipage}
    \begin{minipage}[c]{2.1in}
        \centering
        \subfloat[]{
            \label{fig:irsu_ped_tracking_results:b} 
            \includegraphics[width=2.0in]{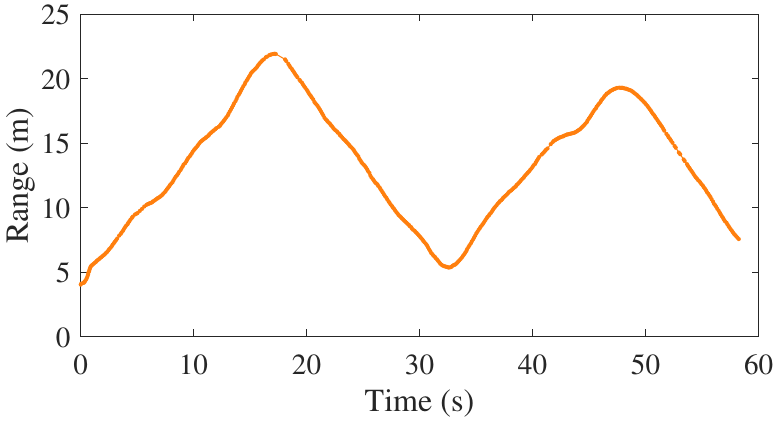}}
        \vfill
        \subfloat[]{
            \label{fig:irsu_ped_tracking_results:c} 
            \includegraphics[width=2.0in]{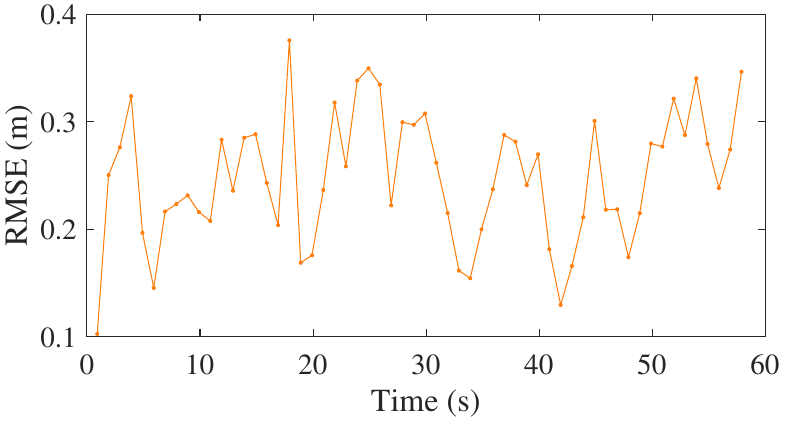}}
    \end{minipage}
    \caption{Pedestrian tracking results for the IRSU. It can be seen in (a) that when the target pedestrian was walking in a figure eight pattern in front of the IRSU the tracked positions are found to be well aligned with the ground truth path obtained from a RTK receiver.}
    \label{fig:irsu_ped_tracking_results} 
\end{figure*}

To assess the position tracking accuracy of the IRSU, an outdoor test was conducted at the Shepherd Street car park at the University of Sydney. Figure \ref{fig:irsu_ped_tracking_setup} illustrates the setup of the test at the car park, where a pedestrian was walking in front of the IRSU for approximately one minute. The ground truth positions of the target pedestrian were obtained at 1 Hz by an u-blox C94-M8P RTK receiver, which reports a standard deviation of 1.4 cm of positioning when in the RTK fixed mode.

As presented in Figure \ref{fig:irsu_ped_tracking_results}, the trajectory of the target reported by the local tracker is found close to the ground truth points. The root mean squared error (RMSE) in position is calculated by comparing the ground truth positions and the corresponding estimates from the tracker. As the two sources of positions were obtained at different rates (10 Hz from the tracker versus 1 Hz from RTK), each RTK reading is compared with the position estimate that has the nearest timestamp. It can be seen from Figure \ref{fig:irsu_ped_tracking_results:c} that the distance of the target pedestrian to the IRSU varies from 5 to 22 metres, and the position RMSE values remain less than 0.4 m throughout the test.

\subsection{Connected and Automated Vehicle}
\label{sec:cav}

\begin{figure*}[!t]
	\centering
	\includegraphics[width=5.0in]{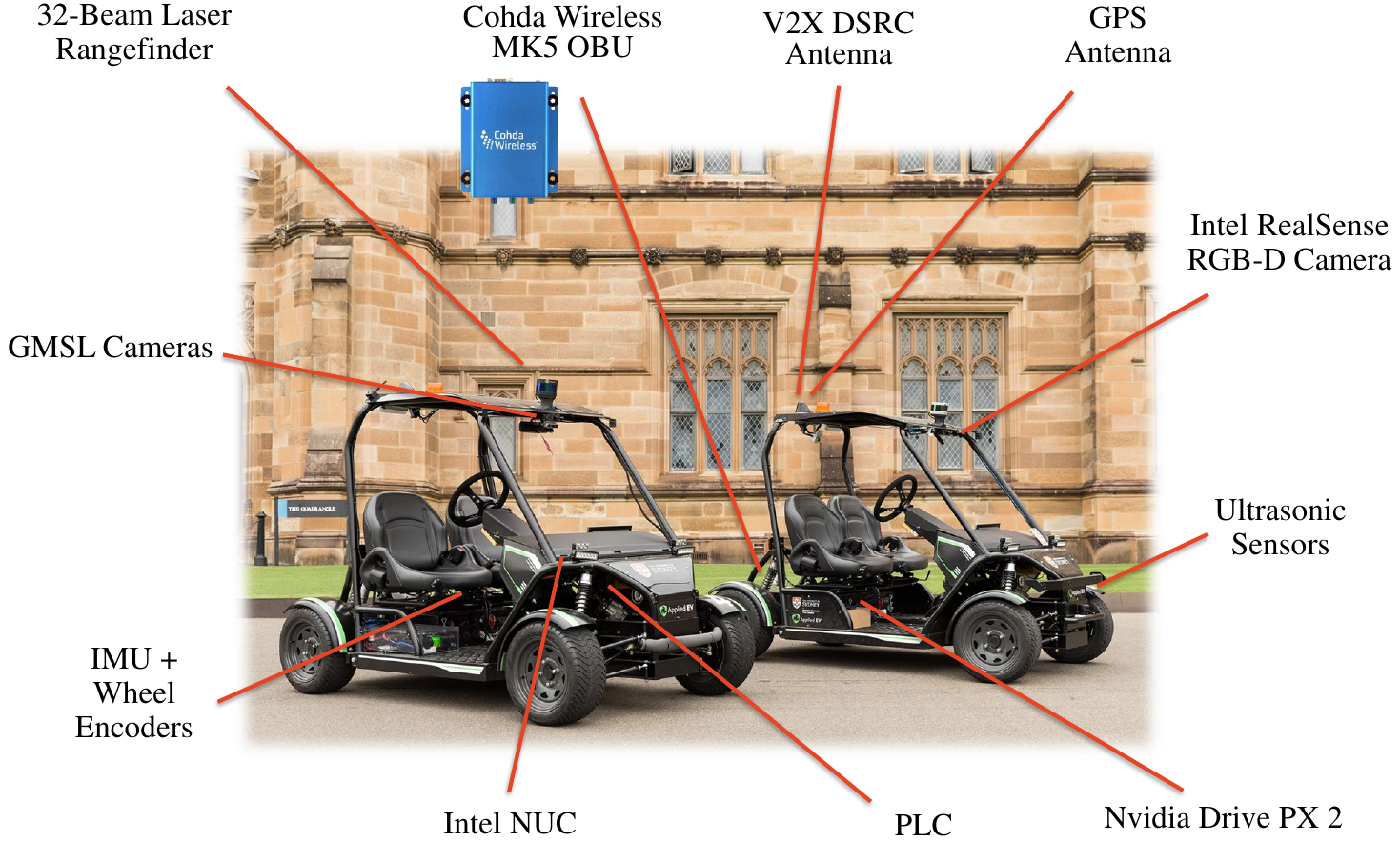}
	\caption{The CAV platform and onboard sensors.}
	\label{fig:cav_platform}
\end{figure*}

Figure \ref{fig:cav_platform} presents an overview of hardware configuration of the CAV platform built by the ACFR ITS group. Images are captured onboard at 30 FPS by an NVIDIA Drive PX2 automotive computer with six gigabit multimedia serial link (GMSL) cameras with 1080p resolution and a \(100^{\circ}\) horizontal FOV each. They are arranged to cover a \(360^{\circ}\) horizontal FOV around the vehicle. The vehicle also has a 32-beam scanning lidar with \(30^{\circ}\) vertical FOV and \(360^{\circ}\) horizontal FOV for scanning the surrounding at 10 Hz. Both the cameras and the lidar have been calibrated to the local coordinate system of the platform, using the automatic extrinsic calibration toolkit presented in \cite{paper:VermaBerrio2019}. Besides, the vehicle has a GNSS receiver, a 6 degrees of freedom IMU, and four wheel encoders for odometry and localisation. The onboard Intel next unit of computing (NUC) has 32 GB of memory and a quad-core Intel i7-6670HQ processor, serving as the main processing computer within the CAV. The NUC is running ROS Melodic on Ubuntu 18.04.2 LTS. Last but not least, the CAV platform has been retrofitted with Cohda Wireless MK5 OBU to enable the V2X communication capability. Please refer to \cite{paper:ZhouBerrio2020} for more details on the CAV platform and the USyd Campus data set collected using the platform.

The vehicle did not use any of the retrofitted perception sensors for road user detection in the simulation or in experiments presented in the paper. The multiple cameras were used for video recording purpose, and the multibeam lidar was enabled only for aiding self-localisation within the map. Lidar feature maps of the experiment sites were built using a simultaneous localisation and mapping (SLAM) algorithm. The maps are based on pole and building corner features extracted from lidar point clouds, which are essential for localisation since GNSS cannot provide the desired level of accuracy in the experiment environments. Interested reader can refer to \cite{paper:YiWorrall2019} for more information. In the meantime, a \textit{Lanelet2} map is built for every experiment site, which includes road network, lane layout and traffic rules such as speed limits, traffic lights and right-of-way rules.

When the CAV receives an ETSI CPM through the onboard Cohda Wireless MK5 OBU, the received perceived objects information is first decoded from binary ASN.1 encoding, and transformed with its uncertainty to the local frame of reference of the CAV, as presented in Section \ref{sec:transformation}, which also takes into account the estimated egocentric pose of the CAV in self-localisation.

Following the coordinate transformation with uncertainty, the perceived objects information from the IRSU are fused into a multiple road user tracking algorithm that is a variant of GMPHD filter running within the local frame of the receiving CAV. The tracked states of road users include their position, heading, and speed. The general formulation of the GMPHD tracker is to use Gaussian mixture to represent the joint distribution of the group of tracked targets. The GMPHD approach is considered attractive due to its inherent convenience in handling track initiation, track termination, probabilistic data association, and clutter. Compared with the naive GMPHD algorithm, the road user tracker running in both the IRSU and CAV is improved with measurement-driven initiation of new tracks, and track identity management. Also note that an instance of the tracker is required for each type of road users, which effectively reduces the overall computational cost.

The navigation subsystem in the CAV is responsible for path planning, monitoring and controlling the motion of the vehicle from the current position to the goal. A hybrid A* path planner presented in our recent work \cite{paper:NarulaWorrall2020} is running within the CAV to plan a path navigating around obstacles and the tracked pedestrians. It maintains a moving grid-based local cost map that considers 1) structural constraints, such as road and lane boundaries, present in the \textit{Lanelet2} map, 2) obstacles picked up by local perception sensors, and 3) current and future estimates of road users detected and broadcast by IRSU through V2I communication, such that it can plan a smooth, safe, and kinematically feasible path to avoid collision with any other road users the CAV becomes aware of. In the experiments presented in the paper, the use of local perception information in the navigation was minimised or disabled.

\subsection{Handling of ETSI CPMs on IRSU and CAV platforms}
\label{sec:codec}

\begin{figure*}[!t]
	\centering
	\includegraphics[width=6.0in]{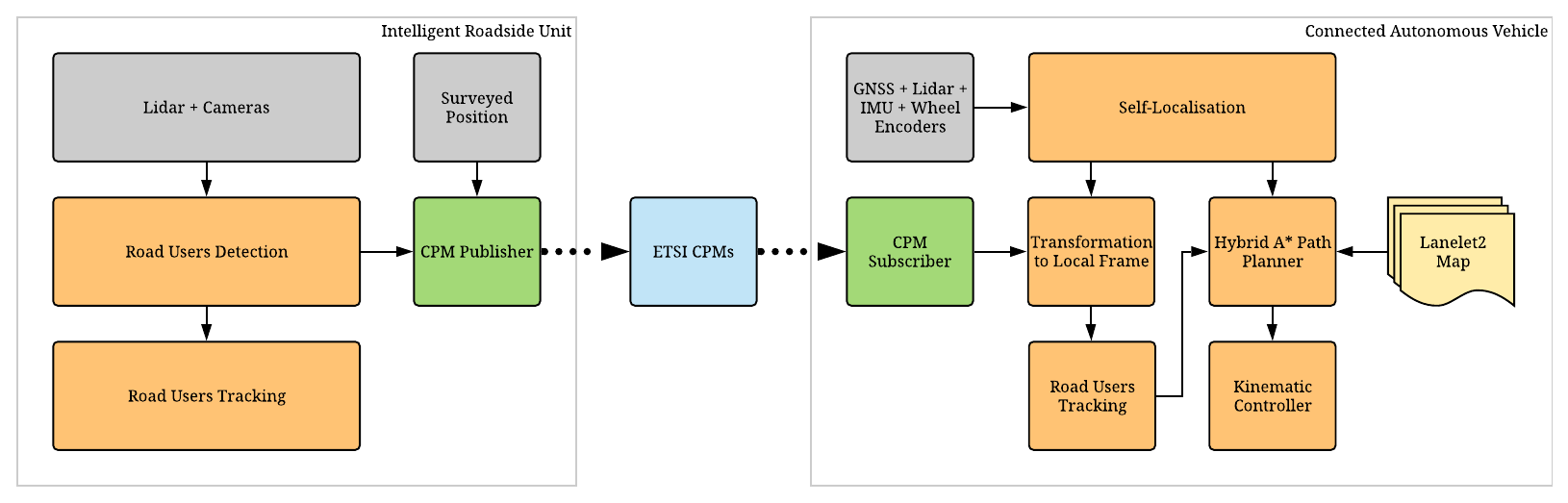}
	\caption{System diagram of the IRSU and the CAV platforms in the experiments.}
	\label{fig:system_structure}
\end{figure*}

\begin{figure*}[!t]
	\centering
	\includegraphics[width=6.0in]{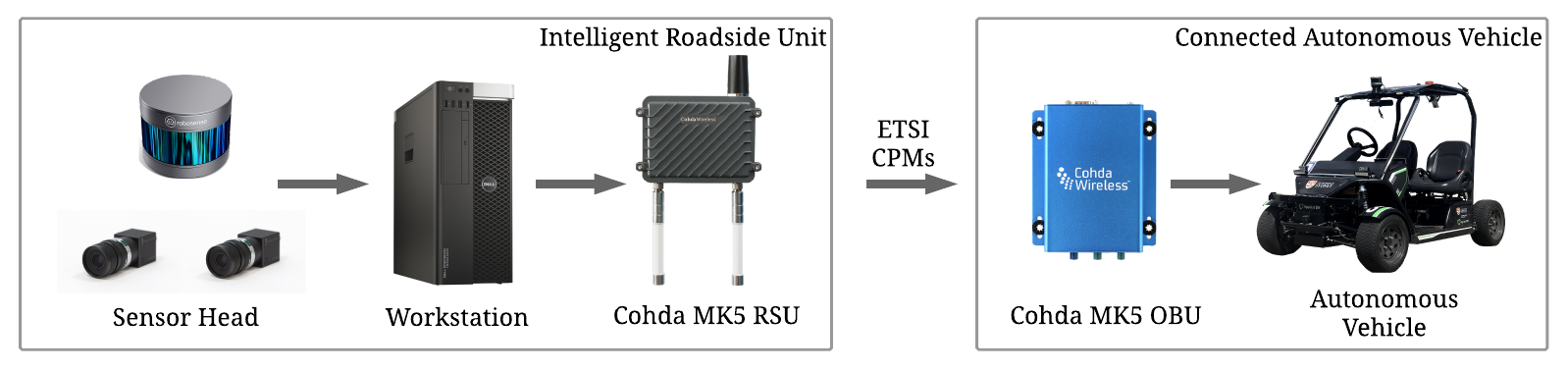}
	\caption{Perception information flow from the IRSU to the CAV. The IRSU broadcasts perceived objects information in the form of ETSI CPMs through Cohda Wireless MK5 RSU. As the receiving agent, the CAV has V2X communication capability through Cohda Wireless MK5 OBU.}
	\label{fig:info_flow}
\end{figure*}

The overall system diagram of the IRSU and the CAV in a C-ITS configuration is illustrated in Figure \ref{fig:system_structure}. The detected road user descriptions are encoded to ETSI CPMs and transmitted from the IRSU to the CAV through the Cohda MK5s at 10 Hz. Each ETSI CPM consists of an ITS PDU header and five types of information containers accommodating mandatory and optional DEs and DFs. These containers are:
\begin{enumerate}
\item A \textit{CPM Management Container}, which indicates station type, such as a vehicle or an IRSU, and reference position of the transmitting ITS-S.
\item An optional \textit{Station Data Container}, which provides additional information about the originating station. This includes the heading, speed, and dimensions when the originating station is a vehicle.
\item Optional \textit{Sensor Information Containers}, which describe the type and specifications of the equipped sensors of the transmitting ITS-S, including sensor IDs, types, and detection areas.
\item Optional \textit{Perceived Object Containers}, each of which describes the dynamics and properties of a perceived object, such as type, position, speed, heading, and dimensions. These perceived object descriptions are registered in the coordinate system of the originating station.
\item Optional \textit{Free Space Addendum Containers}, which describe different confidence levels for certain areas within the sensor detection areas.
\end{enumerate}

The CPM generation rules support different abstraction levels for perceived object descriptions for the implementation flexibility, which can derive from low-level detections made by individual sensors, or the results of the high-level data fusion. In the developed IRSU, the road user detections as a result of camera-lidar fusion are considered in the encoding of CPMs, as depicted in Figure \ref{fig:system_structure}.

The overall perception information flow from the IRSU to CAV is illustrated in Figure \ref{fig:info_flow}. Specifically, the handling of the perception information within the system is mostly carried out in ROS realm. The ROS Kinetic is installed in both Cohda MK5s, with the workstation/NUC set as the ROS master to the transmitting/receiving MK5, respectively. The perceived road users on the workstation side are first described in the form of a ROS CPM, which is used internally within the ROS system and brings convenience of message handling and diagnostics in ROS. Each ROS CPM is a self-contained message that contains all information required for an ETSI CPM. It is the ETSI CPM that is eventually transmitted in the V2X communication.

A ROS \textit{msg\_bridge} node running within each MK5 serves as a bridge between ROS CPMs and ETSI CPMs. It populates an ETSI CPM given information from a ROS CPM and based on the ASN.1 definition of the ETSI CPM. In the meantime, it decodes an ETSI CPM received from other ITS-Ss back into the ROS CPM format. The transmission and reception of ETSI CPM payload is handled by Cohda’s V2X stack in MK5. We keep upgrading the ROS \textit{msg\_bridge} node to support the features required in the CP demonstrations in accordance with the status quo of ETSI CPM standardisation.

\section{Coordinate Transformation of Perceived Objects with Uncertainty}
\label{sec:transformation}

\begin{figure*}[!t]
	\centering
	\includegraphics[width=3.5in]{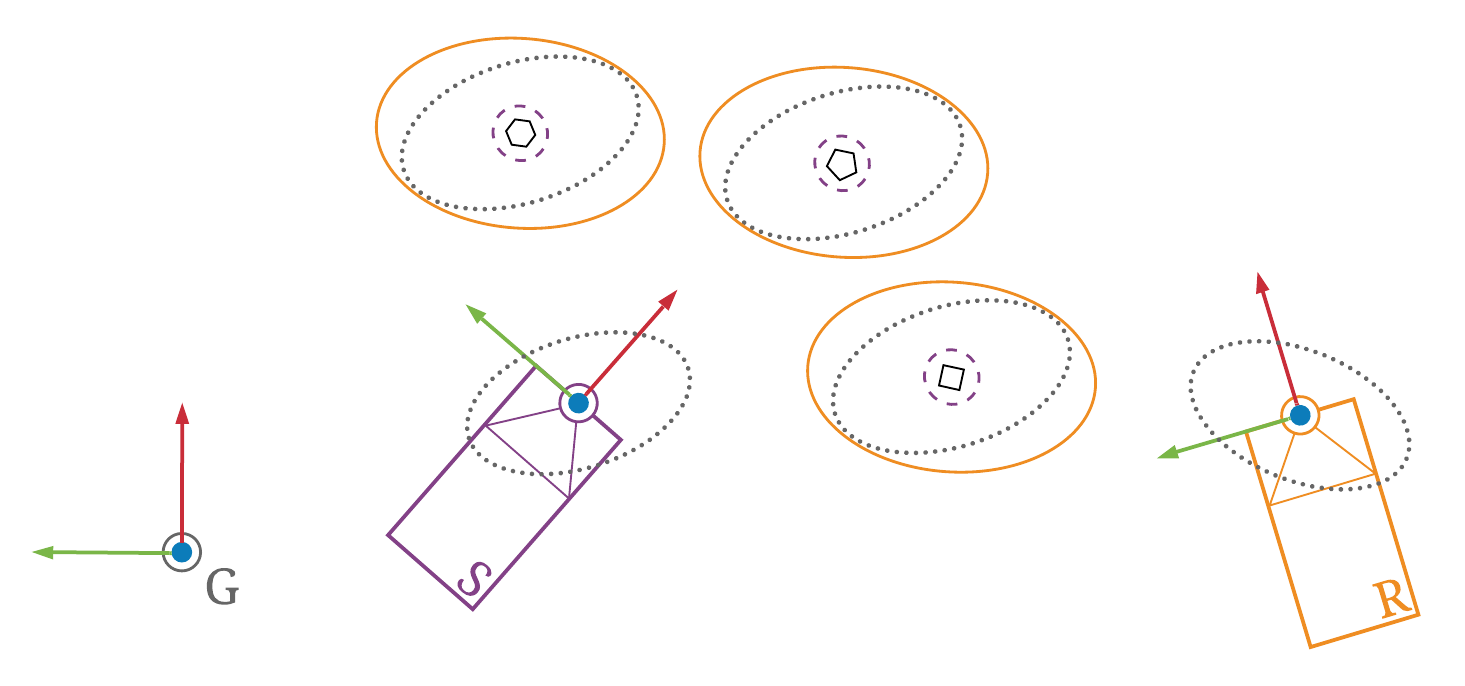}
	\caption{Coordinate transformation of perception information with uncertainty considered from the sensing ITS-S  $S$ to the receiving ITS-S $R$.}
	\label{fig:Transform}
\end{figure*}

The perceived objects information broadcast through CP service is produced from the perspective of the sensing ITS-S. When a piece of perceived object information is received, it is not usable to the receiving agent until it is transformed into its local coordinate system. In \cite{paper:GuntherMennenga2016}, the coordinate transformation of perceived objects does not explicitly incorporate the accuracy fields accompanied with DEs in the EPMs. We stress that uncertainty bounds associated with the perception information are indispensable in successive data fusion and thus have to be considered. Besides, it is essential for the coordinate transformation to also incorporate the uncertainty contained in the pose estimation of both ITS-Ss. These are discussed in detail as follows.

\begin{enumerate}
\item Perception uncertainty in the sensing ITS-S. Every commonly used perception sensor in ITS area has its own strengths and limitation. For instance, RGB images are useful in detecting object instances and classification of road users with an estimated confidence level using a visual classifier algorithm. Doppler RADARs produce both position and velocity measurements but are prone to noise and interference from the environment. Lidars have a high range resolution to observe physical extent and shape of objects. Nevertheless, the point density decreases dramatically along with range. These sensors all produce measurements corrupted by noise, and thus should be modelled with uncertainty. Combining multi-modal sensory information not only improves robustness of the perception system, but also increases accuracy, which means a lower level of uncertainty. The produced estimates of perceived objects with uncertainty have been represented in \textit{Perceived Object Container} of the CPM specification.
\item Self-localisation uncertainty of sensing and receiving ITS-Ss. A vehicle ITS-S has to constantly localise itself within a global frame of reference such as map or UTM frame for navigation and safety reasons. However, accurate self-localisation for a moving platform is known to be one of the existing challenges in ITS applications such as urban navigation. Using GNSS as the only source for localisation often cannot achieve satisfactory accuracy, in particular in GNSS-degraded or even GNSS-denied environments. There are various existing solutions that can provide higher localisation accuracy such as RTK, GNSS and inertial/encoder data fusion, feature based localisation based on existing map, etc. Nevertheless, even with the same localisation approach, the uncertainty magnitude in the localisation can vary significantly depending on certain external conditions, such as GNSS satellite visibility in the sky, and the quantity of observable features and their qualities in the surroundings. The location of an IRSU, although deployed static in a traffic environment, is not immune from localisation error either when set either by GNSS or through a surveying process. The localisation uncertainties of both sensing and receiving ITS-Ss therefore have to be considered in the perceived objects coordinate transformation since it cannot be completely eliminated regardless of the self-localisation means employed. The originating ITS-S information including its pose with associated uncertainty has been contained within the \textit{CPM Management Container} and \textit{Station Data Container} in the CPM definition.
\end{enumerate}

Consequently, the transformed object state estimates contain uncertainty that is mainly a combined result of uncertainty in the localisation of both ITS-Ss and that in the sensory perception. As illustrated in Figure \ref{fig:Transform}, both sensing and receiving ITS-Ss contain uncertainties in their estimated egocentric states within the global frame $G$. The sensing ITS-S $S$ observes multiple road users within its local frame with some level of sensory measurement uncertainty labelled with purple dashed lines. The road users measurement uncertainties grow when transformed to a global frame $G$ due to self-localisation uncertainty of ITS-S $S$, as labelled with gray dotted lines. The road users measurement uncertainties further grow, as labelled by orange solid lines, when transformed to the local frame of the receiving ITS-S $R$ as its self-localisation uncertainty is also incorporated.

Both the work in \cite{paper:RauchKlanner2012} and \cite{paper:AlligWanielik2019IV} performs the coordinate transformation of perceived object states with their uncertainty through temporal and spatial alignment. The temporal alignment is introduced in the work mainly due to the time difference between the reception of a CAM and a Ko-PER CPM. It is however less of a concern for an ETSI CPM, which now can contain the originating station information required from a CAM.


\subsection{Problem Formulation}

This section presents mathematical formulation of the coordinate transformation of perceived objects information a recipient ITS-S receives from a sensing ITS-S. We propose to use unscented transform (UT) in the coordinate transformation with uncertainty, which here is recognised as a non-linear and non-deterministic process. The formulation presented is intended for 2D transformation, as it is reasonable to assume a planar road surface for the road user detection and tracking problem discussed in the paper. Nevertheless, the formulation can be easily extended to transformation in 3D space. It is important to note that although the paper is mainly focused on CP service provided by intelligent infrastructure, the formulation presented here is intended to provide a generic form for two arbitrary ITS-Ss in a V2X network, which includes both V2V and V2I scenarios. Before proceeding further, some definitions about coordinate systems are first given to facilitate discussions.

The global frame, which is the fixed coordinate system attached on the ground, is represented as \(\left\{G\right\} = \left\{\overrightarrow{x}_{G}, \overrightarrow{y}_{G}\right\}\). Local frame of the sensing ITS-S \(S\) is attached on the platform body and is represented as \(\left\{S\right\} = \left\{\overrightarrow{x}_{S}, \overrightarrow{y}_{S}\right\}\), with \(\overrightarrow{x}_{S}\) pointing to the east direction if it is an IRSU and pointing towards its moving direction for a vehicle. Local frame of the recipient ITS-S \(R\) is attached on the platform body and moving with the platform. It is denoted as \(\left\{R\right\} = \left\{\overrightarrow{x}_{R}, \overrightarrow{y}_{R}\right\}\) with \(\overrightarrow{x}_{R}\) pointing towards its moving direction.

The pre-superscript is employed to describe the coordinate frame in which the corresponding variable is expressed. For instance, \({}^{V}\!x_{t}^{p}\) denotes the position of an object \(p\) in x direction with respect to \(\left\{V\right\}\) at time \(t\). A variable without pre-superscript is defined in a generic way to not specify a particular coordinate frame.

The state of the receiving ITS-S \(R\) at time \(t\) and its estimate are denoted as
\begin{equation}
{}^{G}\!\textbf{x}_{t}^{R} = 
\begin{bmatrix}
{}^{G}\!x_{t}^{R} & {}^{G}\!y_{t}^{R} & {}^{G}\!\theta_{t}^{R}\\
\end{bmatrix}
^{T},
\end{equation}
where \({}^{G}\!x_{t}^{R}\), \({}^{G}\!y_{t}^{R}\), and \({}^{G}\!\theta_{t}^{R}\) denote 2D coordinates and the heading, respectively, in \(\left\{G\right\}\) at time \(t\). The state estimate is represented by a multivariate Gaussian as
\begin{equation}
{}^{G}\!\textbf{x}_{t}^{R} \sim \mathcal{N}\left({}^{G}\!\bar{\textbf{x}}_{t}^{R},{}^{G}\!\bm{\Sigma}_{t}^{R}\right),
\end{equation}
where \({}^{G}\!\bar{\textbf{x}}_{t}^{R}\) and \({}^{G}\!\bm{\Sigma}_{t}^{R}\) denote the mean vector and covariance matrix, respectively.

Likewise, the estimated pose of the perceiving ITS-S \(S\) at time \(t\) is too represented as a Gaussian variable:
\begin{equation}
\label{eq:GaussianPerceivingITSS}
{}^{G}\!\textbf{x}_{t}^{S} = 
\begin{bmatrix}
{}^{G}\!x_{t}^{S} & {}^{G}\!y_{t}^{S} & {}^{G}\!\theta_{t}^{S}\\
\end{bmatrix}
^{T}
\sim \mathcal{N}\left({}^{G}\!\bar{\textbf{x}}_{t}^{S},{}^{G}\!\bm{\Sigma}_{t}^{S}\right),
\end{equation}
where \({}^{G}\!x_{t}^{S}\) and \({}^{G}\!y_{t}^{S}\) are the location of \(S\), \({}^{G}\!\theta_{t}^{I}\) is its heading with respect to \(\left\{G\right\}\) at time \(t\), \({}^{G}\!\bar{\textbf{x}}_{t}^{S}\) and \({}^{G}\!\bm{\Sigma}_{t}^{S}\) denote the mean vector and covariance matrix, respectively. In the special case of an IRSU serving as the perceiving ITS-S, the subscription \({t}\) in \eqref{eq:GaussianPerceivingITSS} can be dropped as \({}^{G}\!x_{t}^{S}\) and \({}^{G}\!y_{t}^{S}\) are time invariant and \({}^{G}\!\theta_{t}^{S}\) is towards due east for an IRSU. Nevertheless, \({}^{G}\!\bm{\Sigma}_{t}^{S}\) is still applicable to an IRSU due to the presence of uncertainty in the surveying process.

The state vector of a perceived object \(p\) at time \(t\) is represented from the sensing ITS-S's perspective as
\begin{equation}
{}^{S}\!\textbf{x}_{t}^{p} = 
\begin{bmatrix}
{}^{S}\!x_{t}^{p} & {}^{S}\!y_{t}^{p} & {}^{S}\!\theta_{t}^{p}\\
\end{bmatrix}
^{T}
\sim \mathcal{N}\left({}^{S}\!\bar{\textbf{x}}_{t}^{p},{}^{S}\!\bm{\Sigma}_{t}^{p}\right),
\end{equation}
where \({}^{S}\!x_{t}^{p}\) and \({}^{S}\!y_{t}^{p}\) are object coordinates and \({}^{S}\!\theta_{t}^{p}\) is heading within the sensing ITS-S's local frame at time \(t\).

We obtain the transformed state of perceived object \(p\) with respect to \(\left\{R\right\}\) by
\begin{equation}
\label{eq:Transform}
{}^{R}\!\textbf{x}_{t}^{p} = 
trans\left({}^{G}\!\textbf{x}_{t}^{R}, {}^{G}\!\textbf{x}_{t}^{S}, {}^{S}\!\textbf{x}_{t}^{p}\right).
\end{equation}

Specifically, in \(trans\left(\cdot\right)\), we have
\begin{equation}
\begin{split}
\begin{bmatrix}
{}^{R}\!x_{t}^{p}\\
{}^{R}\!y_{t}^{p}\\
1
\end{bmatrix} &= T\left({}^{G}\!\textbf{x}_{t}^{R}\right)^{-1} T\left({}^{G}\!\textbf{x}_{t}^{S}\right)
\begin{bmatrix}
{}^{S}\!x_{t}^{p}\\
{}^{S}\!y_{t}^{p}\\
1
\end{bmatrix}\\
{}^{R}\!\theta_{t}^{p} &=
{}^{S}\!\theta_{t}^{p} + {}^{G}\!\theta_{t}^{S} - {}^{G}\!\theta_{t}^{R}
\end{split},
\end{equation}
where
\(T\left(
\begin{bmatrix}
x & y & \theta\\
\end{bmatrix}
^{T}
\right) =
\begin{bmatrix}
\cos\theta & -\sin\theta & x\\
\sin\theta & \cos\theta & y\\
0 & 0 & 1
\end{bmatrix}
\)
is the homogeneous transformation matrix.

Given a Gaussian representation of perceived object state pdf in the form of mean vector \(\bar{\textbf{x}}\) and covariance matrix \(\bm{\Sigma}\) with respect to \(\left\{S\right\}\), the nonlinear transformation to receiving ITS-S's frame \(\left\{R\right\}\) with uncertainty is achieved through the UT process.

An augmented state vector is constructed by concatenating the state vector of the receiving and sensing ITS-Ss and that of the perceived object \(p\) before transformation. Its Gaussian estimate is written as

\begin{equation}
\label{eq:AugmentedState}
\textbf{x}_{t}^{a} \sim \mathcal{N}\left(\bar{\textbf{x}}_{t}^{a},\bm{\Sigma}_{t}^{a}\right),
\end{equation}
where 
\(
\bar{\textbf{x}}_{t}^{a} = 
\begin{bmatrix}
\left({}^{G}\!\bar{\textbf{x}}_{t}^{R}\right)^{T} & \left({}^{G}\!\bar{\textbf{x}}_{t}^{S}\right)^{T} & \left({}^{S}\!\bar{\textbf{x}}_{t}^{p}\right)^{T}\\
\end{bmatrix}
^{T}
\), and
\(
\bm{\Sigma}_{t}^{a} = 
\text{blkdiag}\left\{{}^{G}\!\bm{\Sigma}_{t}^{R}, {}^{G}\!\bm{\Sigma}_{t}^{S}, {}^{S}\!\bm{\Sigma}_{t}^{p}\right\}.
\)

\begin{figure*}[!t]
	\centering
	\includegraphics[width=4.5in]{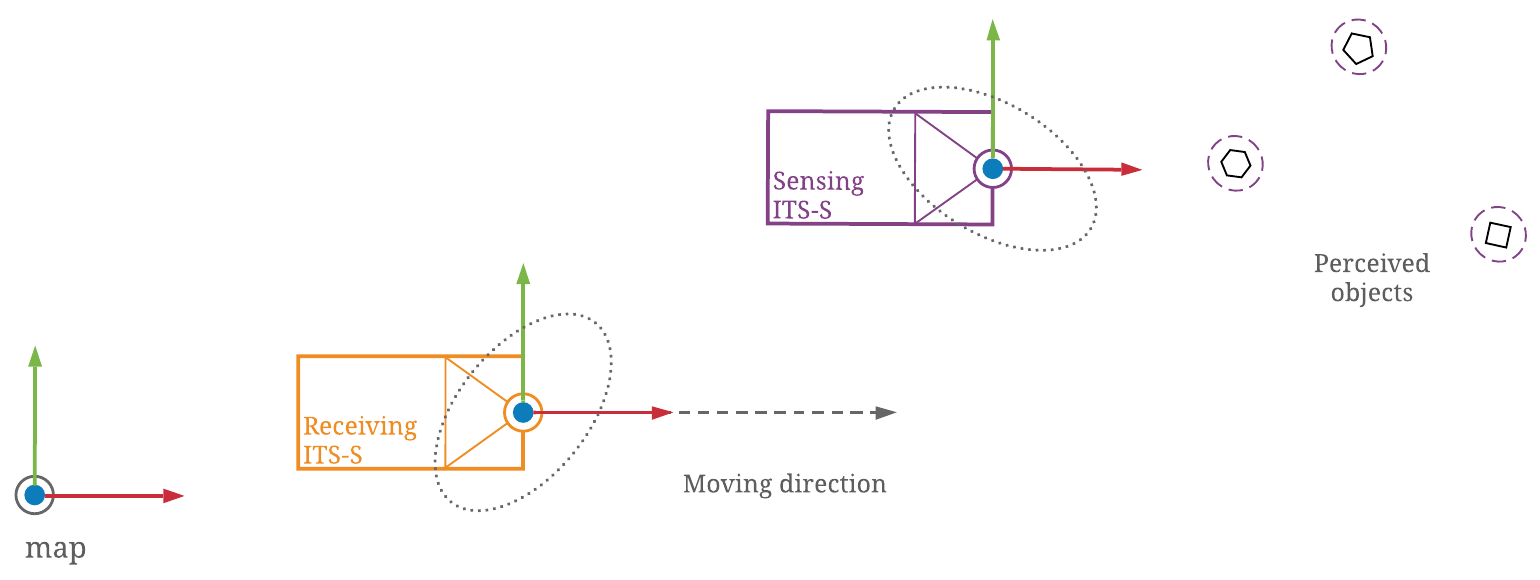}
	\caption{Setup of the sensing and receiving ITS-Ss in the simulation.}
	\label{fig:sim_setup}
\end{figure*}

\begin{figure*}[!t]
	\centering
	\includegraphics[width=4.0in]{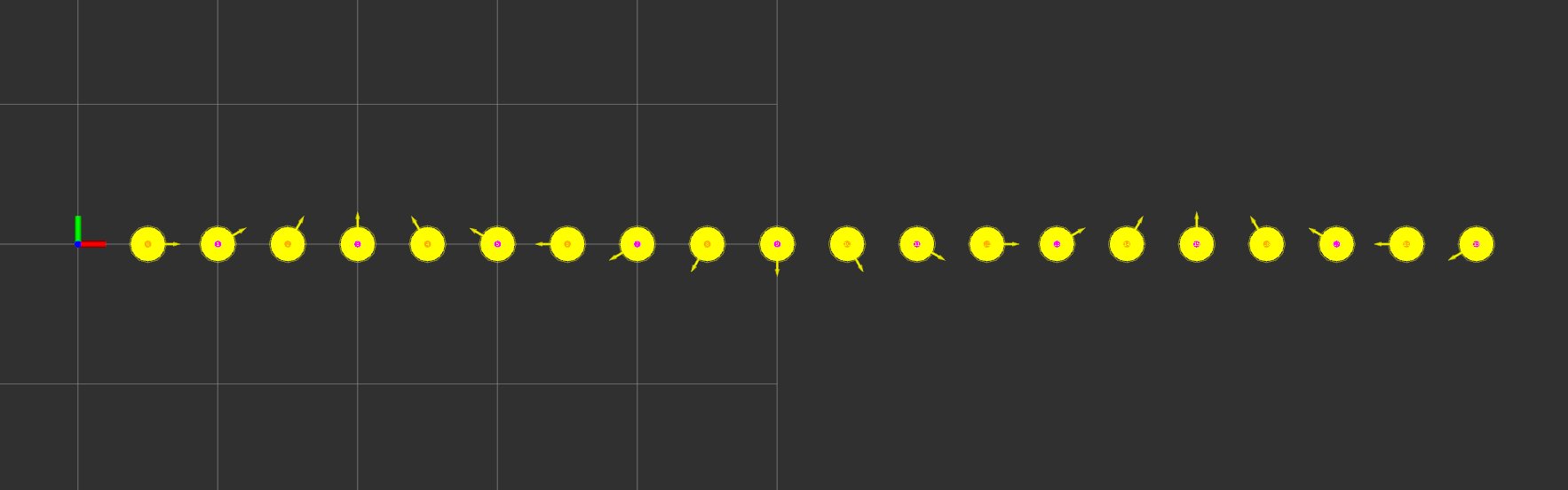}
	\caption{Perceived objects within the local frame of the sensing ITS-S. The objects are a mixture of static pedestrians and vehicles that are placed in a line along with the $x$ direction of the sensing ITS-S. They are perceived with the same location and heading uncertainties in the simulation. The 95\% confidence ellipse for the 2D position estimate of each perceived object is shown in yellow. Each grid in every figure represents an area of $10 \times 10 m$.}
	\label{fig:sim_perceived_objects}
\end{figure*}

\begin{table*}[!t]
\caption{Simulation Parameters.}
\centering
\begin{tabular}{cccc}
\toprule
\multicolumn{2}{c}{\multirow{2}{*}{}} & \multicolumn{2}{c}{\textbf{Standard Deviation in State Estimate}} \\
\multicolumn{2}{c}{} & Position & Heading \\
\midrule
\multicolumn{2}{c}{Perceived Objects} & $0.5 m$ & $6 ^{\circ}$ \\
\multirow{2}{*}{Sensing ITS-S} & IRSU & $0.005 m$ & $\epsilon$ \\
& CAV & \multicolumn{2}{c}{Same to Receiving CAV} \\
\multirow{2}{*}{Receiving CAV} & Test 1 & $0.25 m$ & $0.05^{\circ}, 0.5^{\circ}, 1.0^{\circ}, 1.5^{\circ}, 2.0^{\circ}$ \\
& Test 2 & $0.005 m, 0.25 m, 0.5 m, 0.75 m, 1.0 m$ & $0.5^{\circ}$ \\
\bottomrule
\end{tabular}
\label{tab:sim_param}
\end{table*}

A collection of sigma points \(\left\{\bm{\mathcal{X}}_{i}, w_{i}^{m}, w_{i}^{c}\right\}_{i=0}^{2d}\) are obtained based on the augmented state estimate prior to the transformation:

\begin{equation}
\begin{split}
\bm{\mathcal{X}}_{0} &= \bar{\textbf{x}}_{t}^{a}\\
\bm{\mathcal{X}}_{i} &= \bar{\textbf{x}}_{t}^{a} + \left(\sqrt{\left(d+\lambda\right)\bm{\Sigma}_{t}^{a}}\right)_i\ \text{for}\ i=1,\cdots,d\\
\bm{\mathcal{X}}_{i} &= \bar{\textbf{x}}_{t}^{a} - \left(\sqrt{\left(d+\lambda\right)\bm{\Sigma}_{t}^{a}}\right)_i\ \text{for}\ i=d+1,\cdots,2d\\
w_{0}^{m} &= \frac{\lambda}{d+\lambda}\\
w_{0}^{c} &= \frac{\lambda}{d+\lambda} + \left(1-\alpha^2+\beta\right)\\
w_{i}^{m} &= w_{i}^{c} = \frac{1}{2\left(d+\lambda\right)}\ \text{for}\ i=1,\cdots,2d
\end{split},
\end{equation}
where \(\lambda = \alpha^2\left(d+\kappa\right)-d\), \(d = dim\left(\textbf{x}\right)\) is the dimension of state \(\textbf{x}\), scaling parameters \(\kappa \ge 0\), \(\alpha \in \left(0, 1\right]\), and \(\beta = 2\) is optimal for Gaussian distributions, \(\left(\sqrt{\bm{\Sigma}_{t}^{a}}\right)_i\) is to obtain the \(i^{th}\) column of the matrix square root \(\textbf{R} = \sqrt{\bm{\Sigma}_{t}^{a}}\), which can be computed by Cholesky decomposition such that we have \(\bm{\Sigma}_{t}^{a} = \textbf{R}\textbf{R}^{T}\). Note that each sigma point can be decomposed in accordance with the concatenation sequence in \eqref{eq:AugmentedState}, i.e., \(\mathcal{X}_{i} = \begin{bmatrix} \left(\mathcal{X}_{i}^{R}\right)^{T} & \left(\mathcal{X}_{i}^{S}\right)^{T} & \left(\mathcal{X}_{i}^{p}\right)^{T} \\ \end{bmatrix}\).

This is followed by passing each sigma point through the frame transformation function \(trans\left(\cdot\right)\) in \eqref{eq:Transform}, which yields a set of transformed sigma points. For \(i=0,1,\cdots,2d\),

\begin{equation}
\bm{\mathcal{Y}}_{i} = 
trans\left(\mathcal{X}_{i}^{R}, \mathcal{X}_{i}^{S}, \mathcal{X}_{i}^{p}\right).
\end{equation}

Lastly, the transformed state of perceived object \(p\) in \(\left\{R\right\}\) is recovered by

\begin{equation}
\begin{split}
{}^{R}\!\bar{\textbf{x}}_{t}^{p} &= \sum_{i=0}^{2d}{w_{i}^{m} \bm{\mathcal{Y}}_{i}}\\
{}^{R}\!\bm{\Sigma}_{t}^{p} &= \sum_{i=0}^{2d}{w_{i}^{c} \left(\bm{\mathcal{Y}}_{i}-{}^{R}\!\bar{\textbf{x}}_{t}^{p}\right) \left(\bm{\mathcal{Y}}_{i}-{}^{R}\!\bar{\textbf{x}}_{t}^{p}\right)^T}
\end{split}.
\end{equation}

\subsection{Numerical Simulation}
\label{sec:num_sim}

The coordinate transformation of perceived objects in ETSI CPMs is validated through numerical simulation. As illustrated in Figure \ref{fig:sim_setup}, the simulation is setup with two ITS-Ss acting as a publishing/sensing ITS-S and a receiving ITS-S of CPMs. The sensing ITS-S is static at a position of $(100 m, 100 m)$ on the map, facing the east direction. It is perceiving road users within its local frame with some level of uncertainty in its sensory measurements. The receiving ITS-S in the simulation is a CAV moving at $2 m/s$ from its initial position at $(0 m, 75 m)$ with the same heading of the sensing ITS-S, which can be either an IRSU (V2I case) or another CAV (V2V case). Both V2V and V2I scenarios are considered in the simulation for a comparison. Both ITS-Ss are assumed to contain uncertainty in their self-localisation.

Within the sensing ITS-S, the perceived objects information and egocentric pose estimate are encoded into CPMs in the form of binary payloads and published at 10 Hz. The receiving CAV, as it moves on the map, decodes the CPMs received and transforms the perceived objects into its local coordinate system. Through coordinate transformation, the uncertainty in the transformed perceived information is a combined result of the sensing uncertainty and the self-positioning uncertainties of both ITS-Ss.

\begin{figure*}[!t]
	\centering
	\subfloat[]{ 
		\label{fig:sim_ori_results:a} 
		\includegraphics[width=3.04in]{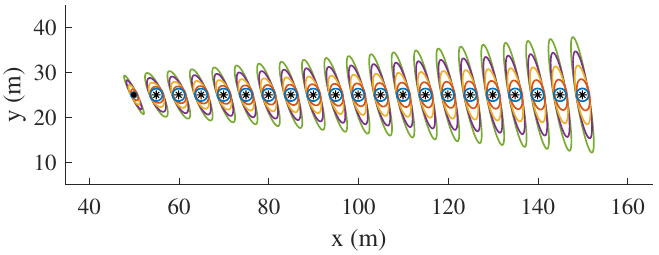}}
	\subfloat[]{ 
		\label{fig:sim_ori_results:b} 
		\includegraphics[width=3.04in]{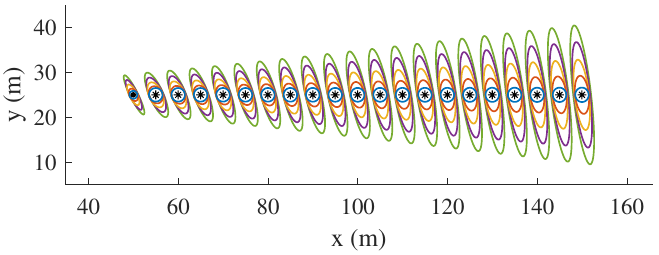}}
	\vfill
	\subfloat[]{ 
		\label{fig:sim_ori_results:c} 
		\includegraphics[width=3.04in]{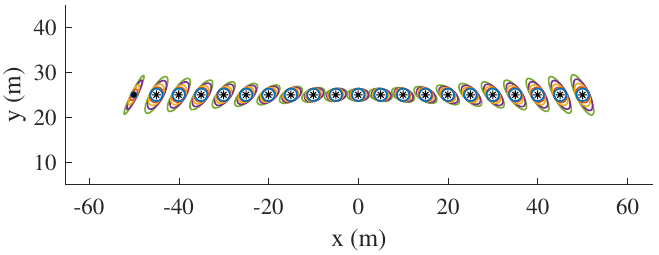}}
	\subfloat[]{ 
		\label{fig:sim_ori_results:d} 
		\includegraphics[width=3.04in]{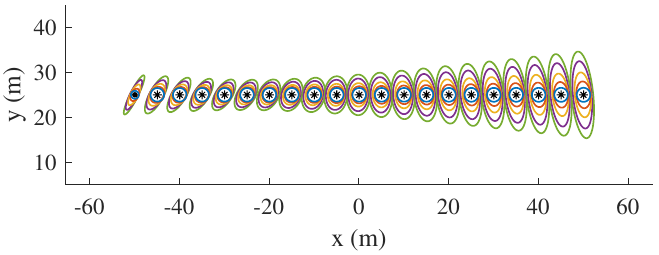}}
	\vfill
	\subfloat[]{ 
		\label{fig:sim_ori_results:e} 
		\includegraphics[width=3.04in]{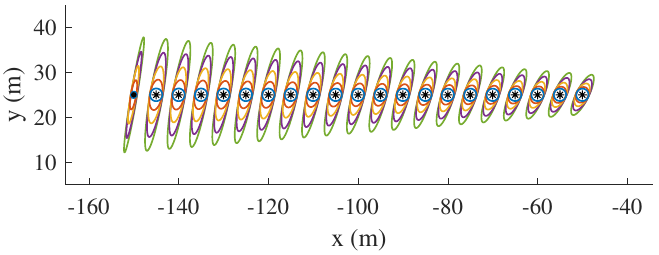}}
	\subfloat[]{ 
		\label{fig:sim_ori_results:f} 
		\includegraphics[width=3.04in]{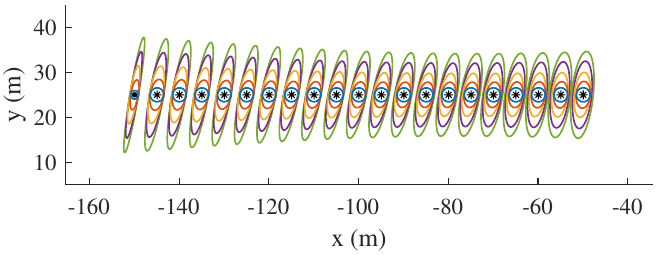}}
	\vfill
	\subfloat[]{ 
		\label{fig:sim_ori_results:g} 
		\includegraphics[width=5.5in]{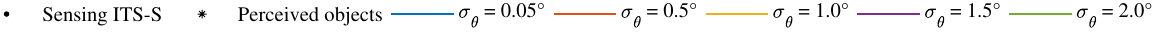}}
	\caption{Confidence ellipses of perceived objects transformed to the local frame of the receiving CAV in \textit{Test 1}, where different heading estimate standard deviations are tested for CAV(s). As the receiving CAV moves, (a), (c), and (e) show the results at different relative positions between IRSU and the receiving CAV, while (b), (d), and (f) present the results for the V2V case. Each ellipse represents 95\% of position estimate confidence. The first ellipse on the left represents the uncertainty in the position estimate of the sensing ITS after transformed into receiving CAV's frame.}
	\label{fig:sim_ori_results} 
\end{figure*}

\begin{figure*}[!t]
	\centering
	\subfloat[]{ 
		\label{fig:sim_pos_results:a} 
		\includegraphics[width=3.04in]{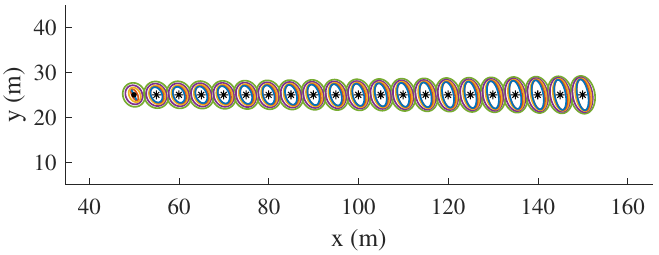}}
	\subfloat[]{ 
		\label{fig:sim_pos_results:b} 
		\includegraphics[width=3.04in]{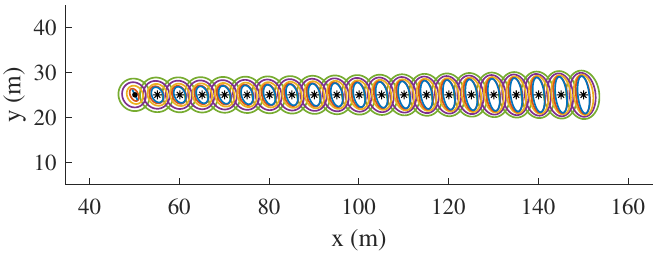}}
	\vfill
	\subfloat[]{ 
		\label{fig:sim_pos_results:c} 
		\includegraphics[width=3.04in]{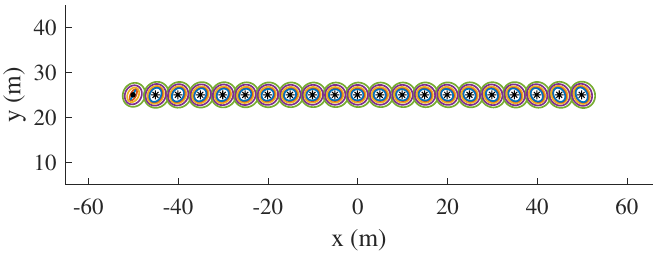}}
	\subfloat[]{ 
		\label{fig:sim_pos_results:d} 
		\includegraphics[width=3.04in]{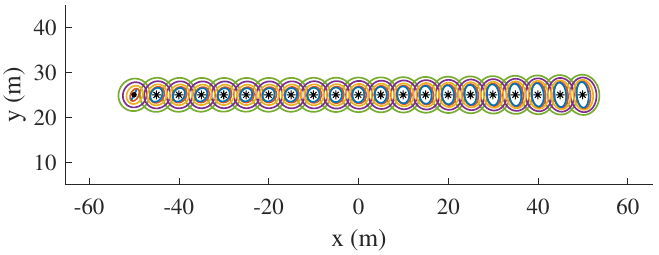}}
	\vfill
	\subfloat[]{ 
		\label{fig:sim_pos_results:e} 
		\includegraphics[width=3.04in]{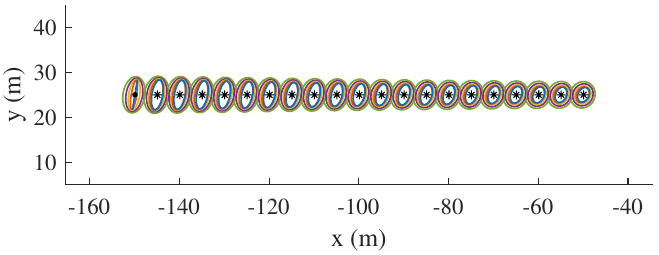}}
	\subfloat[]{ 
		\label{fig:sim_pos_results:f} 
		\includegraphics[width=3.04in]{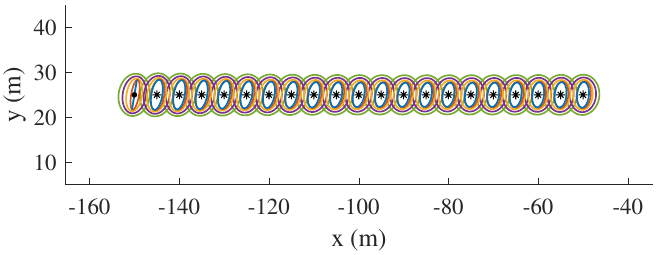}}
	\vfill
	\subfloat[]{ 
		\label{fig:sim_pos_results:g} 
		\includegraphics[width=5.8in]{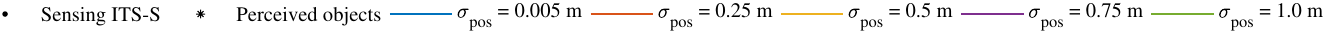}}
	\caption{Confidence ellipses of perceived objects transformed to the local frame of the receiving CAV in \textit{Test 2}, with different standard deviations in position estimate of CAV(s). (a), (c), and (e) demonstrate the results for the V2I case, while (b), (d), and (f) illustrate the results for the V2V case at different relative positions. Each ellipse represents 95\% of position estimate confidence. The first ellipse on the left denotes the uncertainty in the transformed position estimate of the sensing ITS.}
	\label{fig:sim_pos_results} 
\end{figure*}

In the simulation, the covariance matrix in the self-localisation estimate of the receiving CAV is denoted as 
\(
{}^{G}\!\bm{\Sigma}_{t}^{R} =
\text{blkdiag}\left\{\left(\sigma_{pos}\right)^2, \left(\sigma_{pos}\right)^2, \left(\sigma_{\theta}\right)^2\right\}
\), where $\sigma_{pos}$ and $\sigma_{\theta}$ denote the standard deviation of position and heading estimate, respectively. The simulation consists of two tests for evaluating the effect of different $\sigma_{\theta}$ and $\sigma_{pos}$ values in coordinate transformation. Specifically, $\sigma_{\theta}$ is varied in \textit{Test 1} with $\sigma_{pos}$ kept constant, while a range of $\sigma_{pos}$ are tested for the CAV in \textit{Test 2}. Each of the tests further includes both V2V and V2I cases. In the V2V case, the sensing CAV is setup with the same covariance matrix as for the receiving CAV, i.e., \({}^{G}\!\bm{\Sigma}_{t}^{S} = {}^{G}\!\bm{\Sigma}_{t}^{R}\). In the V2I case, where the sensing ITS-S is an IRSU, the localisation noise is assumed small yet non-zero. For simplicity, a group of 20 static road users are positioned in front of the sensing ITS-S in a line with identical perception uncertainty parameters, as demonstrated in Figure \ref{fig:sim_perceived_objects}. Detailed parameters adopted in the simulation can be found in Table \ref{tab:sim_param}. Please note that standard deviations as low as $0.05^{\circ}$ and $0.005 m$ are tested in the simulation as they are roughly the minimum values supported in CPMs for representing the uncertainty of heading and position estimates, respectively, for the originating ITS-S. This is mainly due to discretisation of confidence levels of the corresponding DEs defined in CPMs. Likewise, $0.005 m$ is adopted as the standard deviation for the position estimate of the IRSU, and the heading estimate standard deviation is set to a close-to-zero value $\epsilon$ for preventing numerical errors in the transformation.

Figure \ref{fig:sim_ori_results} reveals the result of \textit{Test 1}, i.e., the effect of different uncertainty levels in heading estimates of the CAV(s), along with the movement of the receiving CAV. Figure \ref{fig:sim_pos_results} depicts the result of \textit{Test 2}, which is with different uncertainty in position estimates of the CAV(s). It can be seen from both figures that the confidence ellipses of those perceived objects, after transformed into the local coordinate system of the receiving CAV, are bloated and distorted to different extents, depending on their relative poses with respect to both ITS-Ss. Please note that in both figures, the first yellow ellipse on the left represents the uncertainty in the transformed position estimate of the sensing ITS.

In Figure \ref{fig:sim_ori_results}, the bloating effect is found sensitive to the uncertainty contained in the heading estimate of both ITS-Ss. Also, the bloated ellipses are slanted along the tangential directions of both ITS-Ss, and the bloating is found more serious for ellipses that are further from both ITS-Ss. In theory, these thin and long confidence ellipses caused by the heading uncertainty are expected to be banana shaped due to the non-linear nature of the coordinate transformation. The result therefore indicates that with a larger uncertainty level in heading estimate and for further perceived objects, the Gaussian assumption starts to show its limitation for representing the perceived object estimates. One can choose to use Gaussian mixture or non-Gaussian representations in the transformation to alleviate the issue. As a comparison, the bloating effect as a result of the uncertainty in position estimates of the ITS-Ss is found less correlated to the relative distance, and the bloating happens in both $x$ and $y$ directions, as shown in Figure \ref{fig:sim_pos_results}. This shows that reducing the estimate uncertainty of heading in ITS-S self-localisation is more effective than that of position, in suppressing the bloating of uncertainty during the coordinate transformation of perceived objects information.

Also, the two scenarios of an IRSU and a CAV acting as the sensing ITS-S are compared in each of Figure \ref{fig:sim_ori_results} and Figure \ref{fig:sim_pos_results}. When the sensing ITS-S is an IRSU, it can be seen that the transformed confidence ellipses are still slanted but in general less bloated due to very low uncertainty in IRSU’s position.

\section{Demonstrations}
\label{sec:results}

\subsection{Experiment in An Urban Traffic Environment}

\begin{figure*}[!t]
	\centering
	\subfloat[]{ 
		\label{fig:abercrombie_setup:a} 
		\includegraphics[width=3.0in]{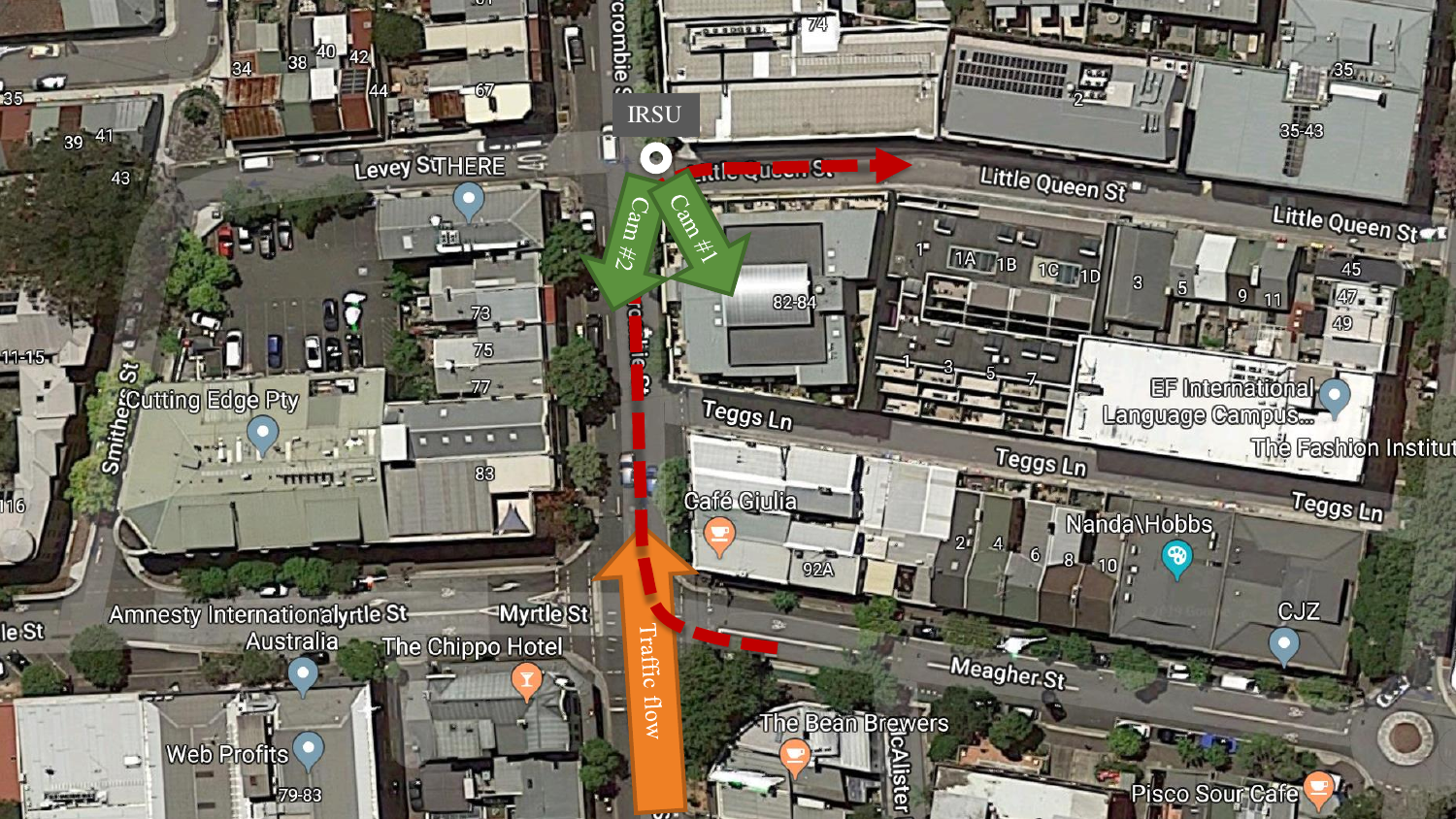}}
	\subfloat[]{ 
		\label{fig:abercrombie_setup:b} 
		\includegraphics[width=3.0in]{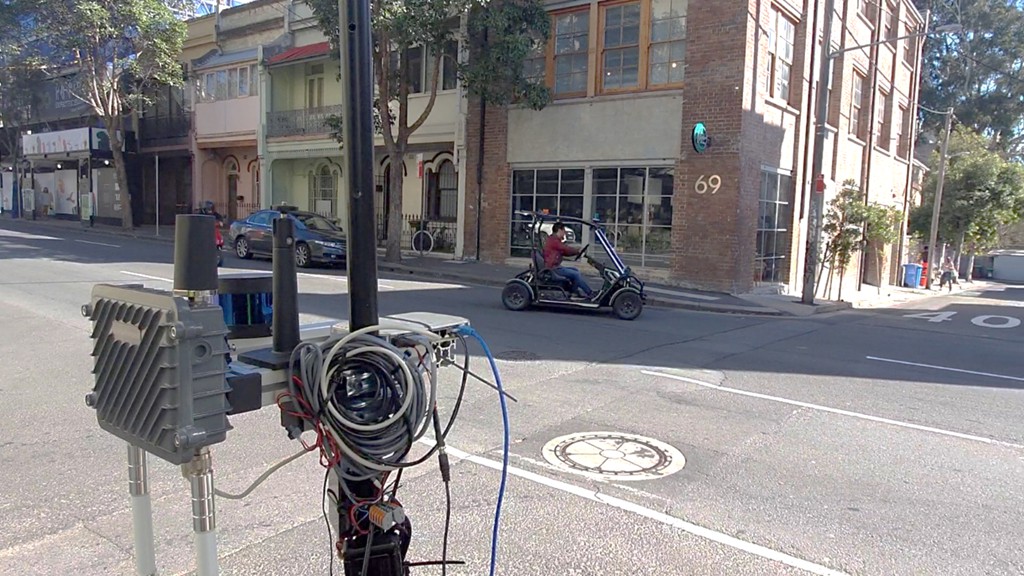}}
	\caption{Experiment setup at the intersection of Abercrombie St and Little Queen St. In (a), the IRSU was setup facing south of Abercrombie St. The car traffic flow was going north on this one-way street. The trajectory of the CV is denoted by a dashed red line, which indicates that the CV turned from Meagher St to Abercrombie St, followed by Little Queen St. In (b), CPMs were transmitted from the IRSU to the moving CV through Cohda MK5s.}
	\label{fig:abercrombie_setup} 
\end{figure*}

\begin{figure*}[!t]
	\centering
	\subfloat[]{ 
		\label{fig:abercrombie_irsu:a} 
		\includegraphics[width=2.8in]{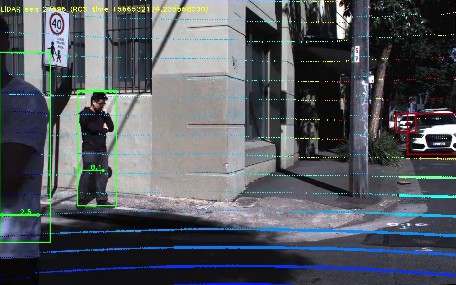}}
	\subfloat[]{ 
		\label{fig:abercrombie_irsu:b} 
		\includegraphics[width=2.8in]{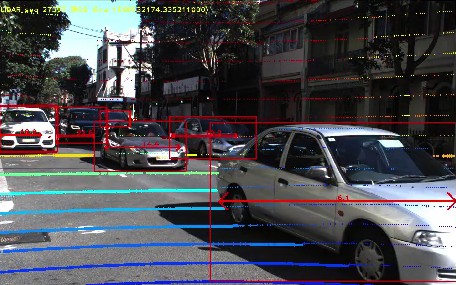}}
	\vfill
	\subfloat[]{ 
		\label{fig:abercrombie_irsu:c} 
		\includegraphics[width=2.8in]{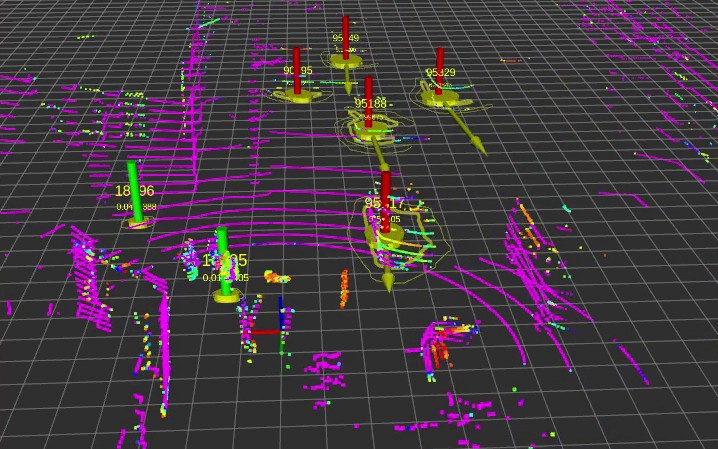}}
	\subfloat[]{ 
		\label{fig:abercrombie_irsu:d} 
		\includegraphics[width=2.8in]{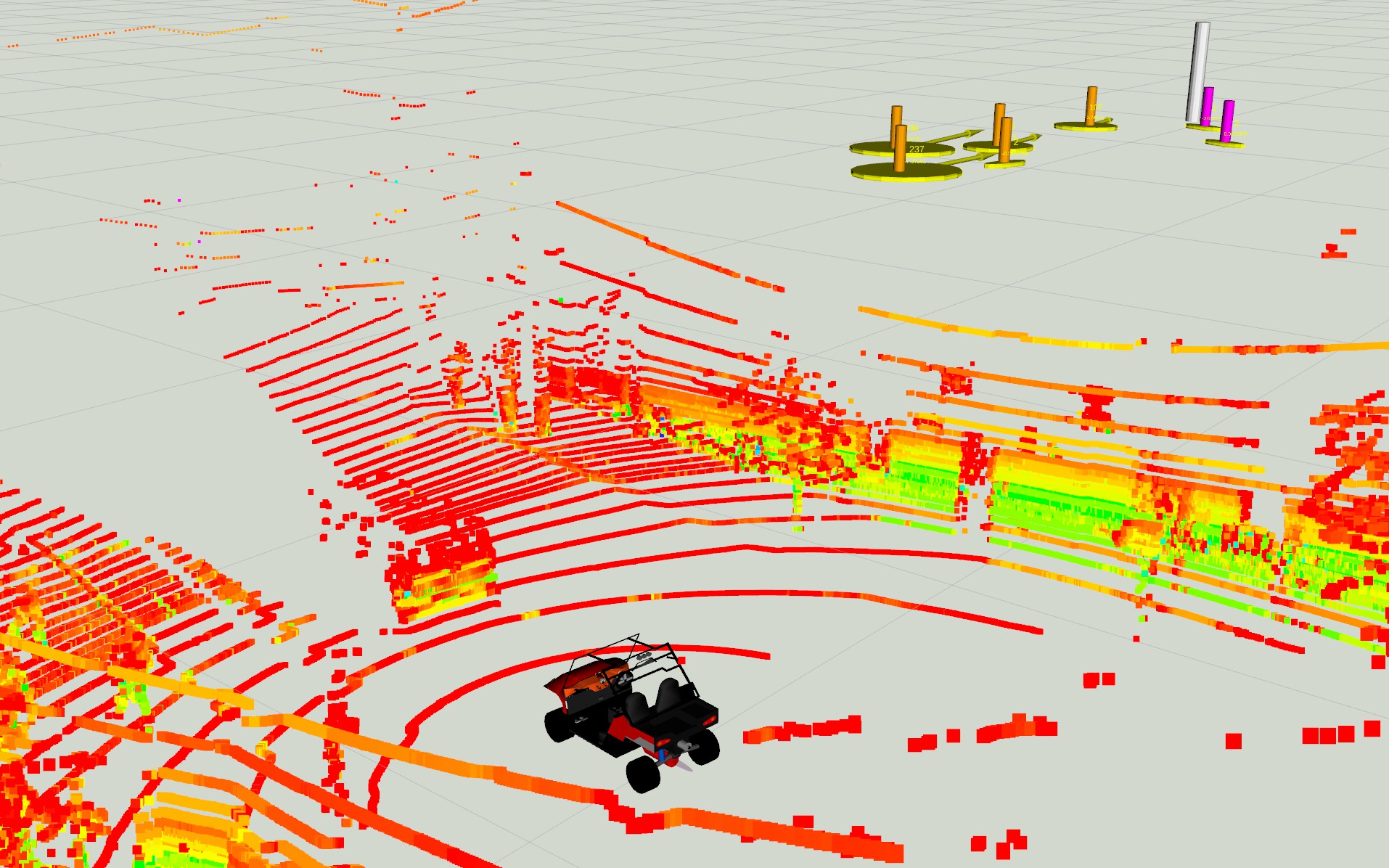}}
	\caption{The road user detection and tracking in the IRSU and CV in the Abercrombie St experiment. (a) and (b) show the image frames from Cam \#1 and \#2, respectively, overlaid with pedestrians and vehicles bounding boxes and projected lidar point cloud in the IRSU. The detection results from the dual cameras and the lidar were then fused within the IRSU for tracking in 3D space, as visualised in (c). The detection results were also transmitted to the CV, where they were transformed and used from the CV's local frame. (d) illustrates the tracking results of the same group of road users within the CV at the same time. Due to change of visualisation colours in IRSU and CV, the vehicles in orange pillars in (d) correspond to red ones in (c), while the pedestrians in magenta pillars correspond to green ones in (c).}
	\label{fig:abercrombie_irsu} 
\end{figure*}

\begin{figure*}[!ht]
	\centering
	\subfloat[]{ 
		\label{fig:abercrombie_results:a} 
		\includegraphics[width=2.5in]{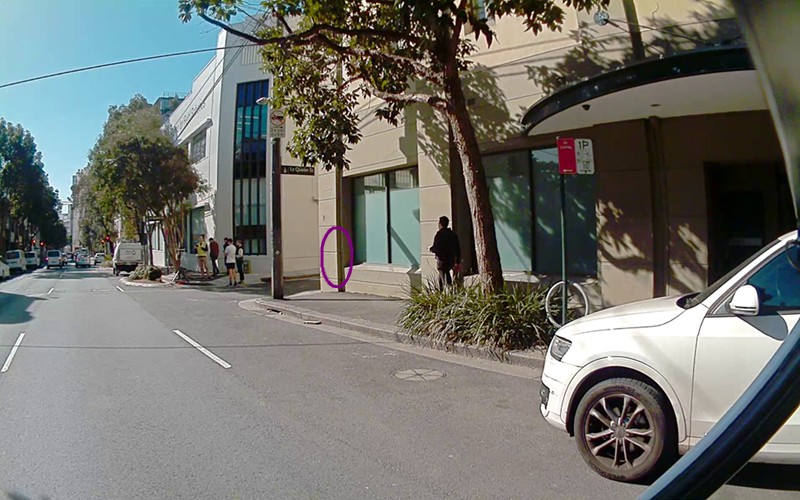}}
	\subfloat[]{ 
		\label{fig:abercrombie_results:b} 
		\includegraphics[width=2.5in]{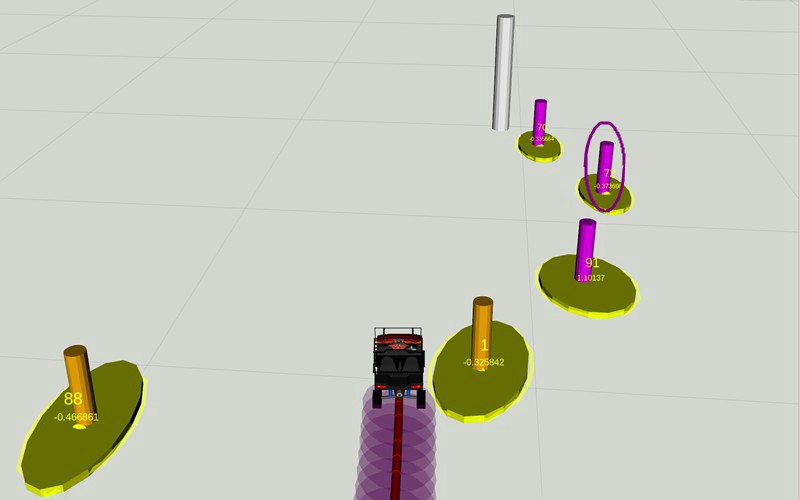}}
	\vfill
	\subfloat[]{ 
		\label{fig:abercrombie_results:c} 
		\includegraphics[width=2.5in]{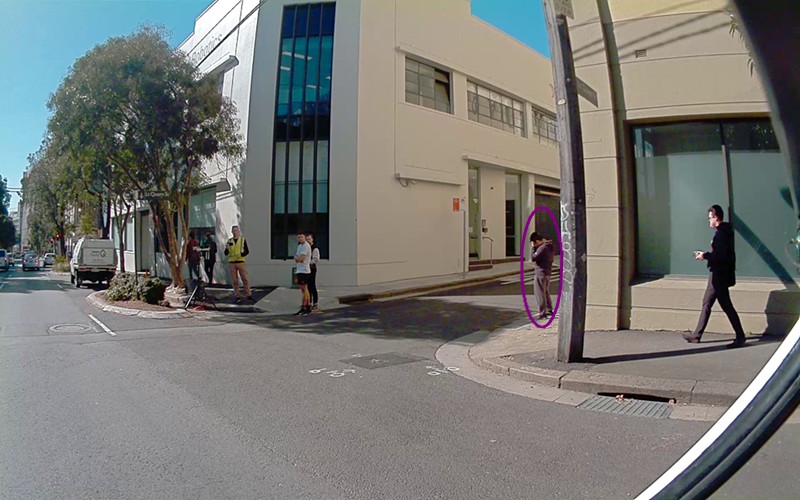}}
	\subfloat[]{ 
		\label{fig:abercrombie_results:d} 
		\includegraphics[width=2.5in]{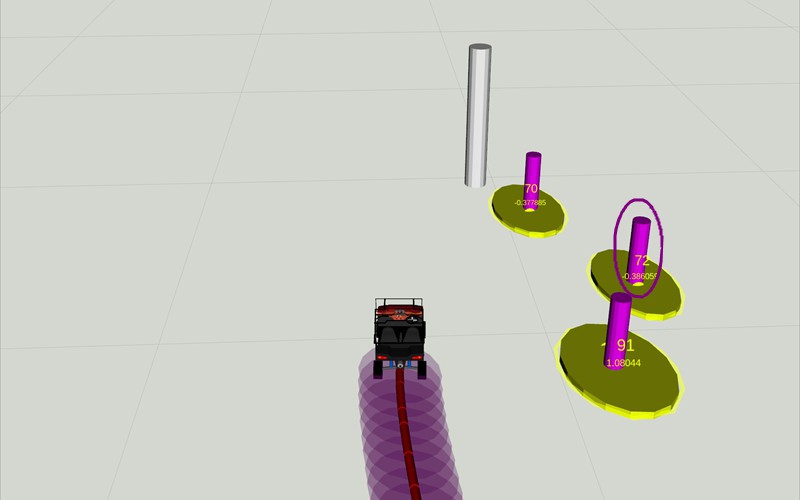}}
	\vfill
	\subfloat[]{ 
		\label{fig:abercrombie_results:e} 
		\includegraphics[width=2.5in]{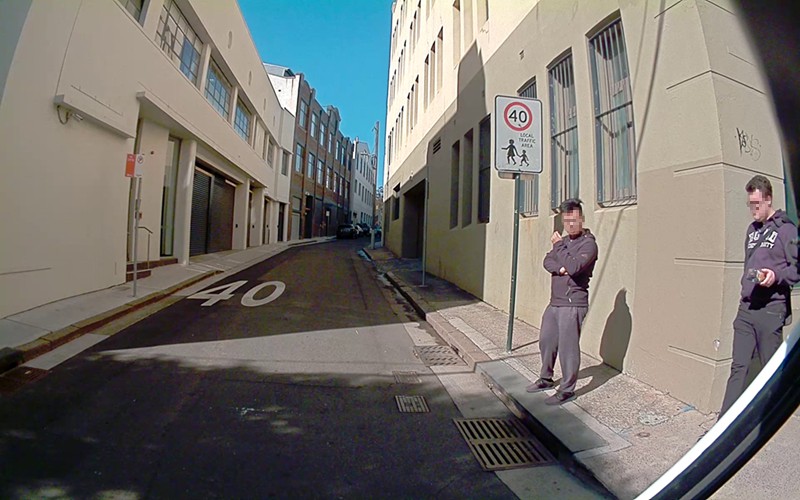}}
	\subfloat[]{ 
		\label{fig:abercrombie_results:f} 
		\includegraphics[width=2.5in]{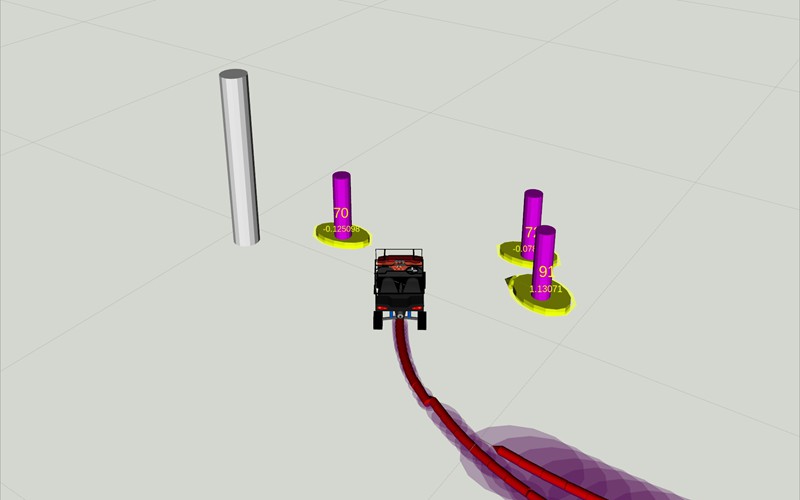}}
	\caption{The CV became aware of a hidden pedestrian before turning from the main street to Little Queen St. The left column shows the images captured by the front-right camera on the CV at different times, the right column illustrates the corresponding road user tracking results within the CV. Also in the right column, the tracked pedestrians are represented as magenta pillars with their 2-sigma estimation confidence areas denoted as yellow ellipses, the IRSU is visualised in the CV as a tall white pillar. In (a) and (b), the CV could “see” through the building a visually occluded pedestrian (labelled with purple circles) around the corner with the help of CPMs received from the IRSU. This is seconds before the CV drove closer and could observe by itself the previously occluded pedestrian on Little Queen St, as depicted in (c) and (d). The localisation covariance is presented as purple disks along the vehicle trajectory in every right column figure. It can be seen in (f) that the uncertainty in road user tracking reduced after the CV received a correction in self-localisation.}
	\label{fig:abercrombie_results} 
\end{figure*}

The experiment was conducted in a real urban traffic environment next to the ACFR building located at Chippendale, Sydney. In the experiment, the IRSU was deployed near the intersection of Abercrombie St and Little Queen St, monitoring the traffic activity at the intersection and broadcasting the perception information in the form of ETSI CPMs in real time, as depicted in Figure \ref{fig:abercrombie_setup}. The intersection was chosen for the experiment mainly because turning from Abercrombie St to Little Queen St, which is a side road, requires extra attention from vehicle drivers due to the lack of traffic control and poor visibility of road users behind the corner building. The main purpose of the experiment is to demonstrate the improved VRU awareness and thus road safety for a vehicle ITS-S when it is able to learn pedestrian activity in blind spots through the CP information provided by IRSU.

Instead of autonomous driving, the CAV in the experiment was driven manually for ensuring public road safety in the real traffic scenario. The CAV platform therefore practically functioned as a CV in the experiment and is referred as the CV in the remainder of the section. While the CV was performing self-localisation within the experiment area, it was able to transform the perceived objects contained in CPMs into its local coordinate system with uncertainty and have the road users tracked in real time. The CV identifies itself and its driver based on the position of the transformed perceived objects in its local frame to avoid self-tracking.

\begin{figure*}[!t]
	\centering
	\includegraphics[width=4.0in]{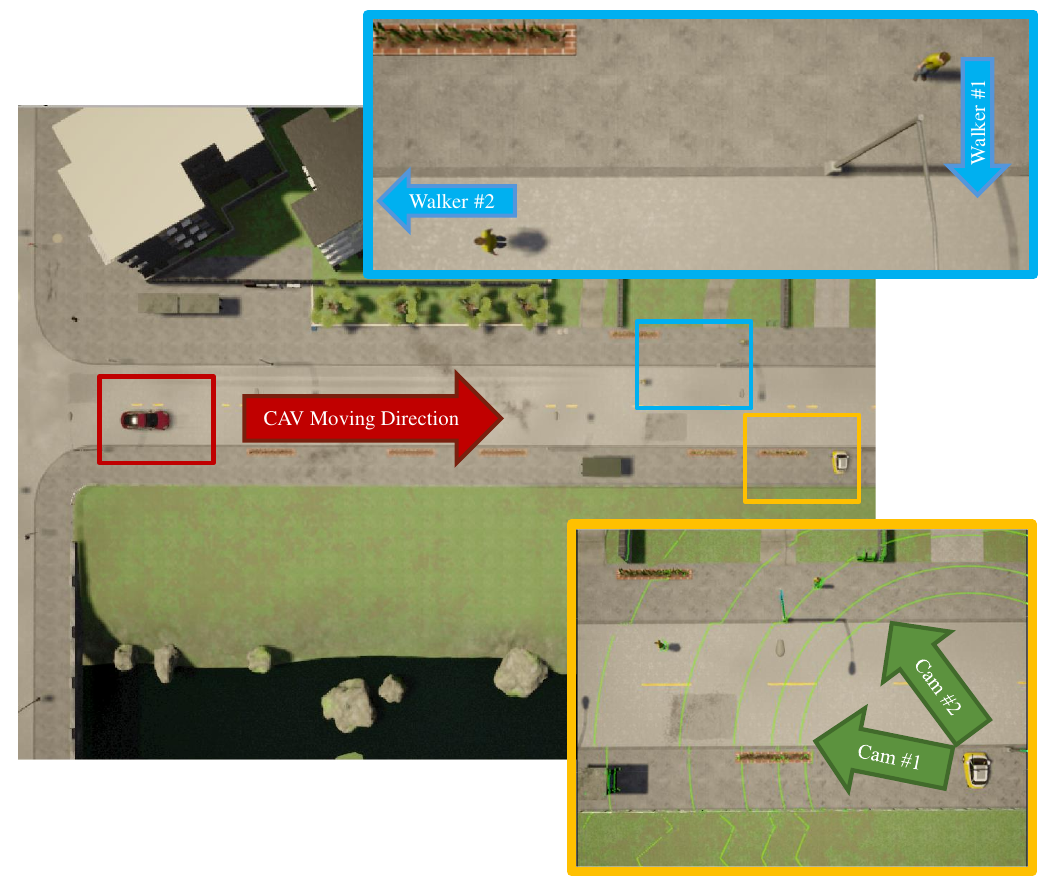}
	\caption{Experiment setup in the CARLA simulator. The CAV, a red car, moved to the east direction towards two walking pedestrians and an IRSU deployed on the roadside. The IRSU was spawn in CARLA as a mini car parked on the roadside.}
	\label{fig:carla_setup}
\end{figure*}

\begin{figure*}[!ht]
	\centering
	\subfloat[]{ 
		\label{fig:carla_irsu:a} 
		\includegraphics[width=2.5in]{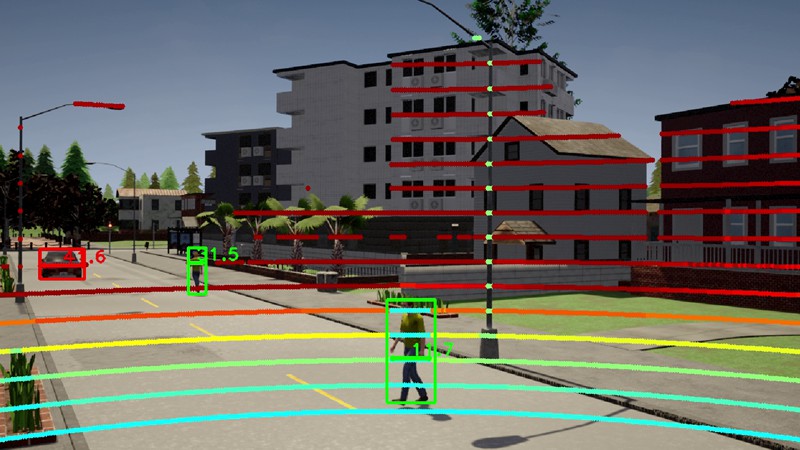}}
	\subfloat[]{ 
		\label{fig:carla_irsu:b} 
		\includegraphics[width=2.5in]{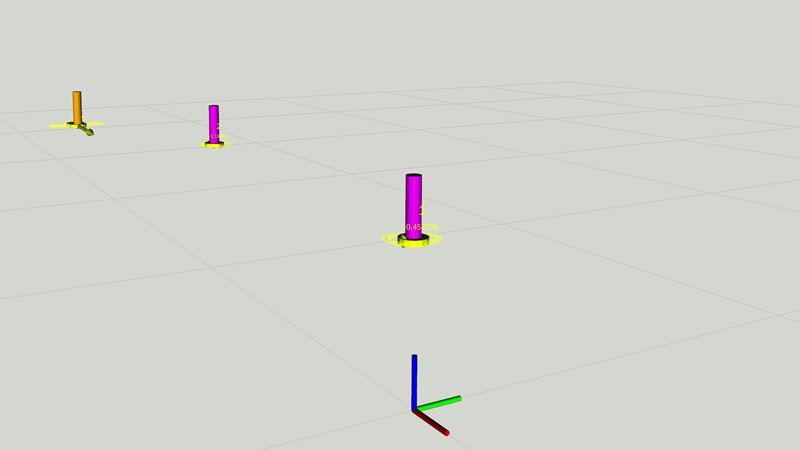}}
	\caption{Pedestrian detection and tracking in the IRSU in the CARLA simulator. Despite dual cameras enabled for detecting road users in the IRSU, the Cam \#1 captured all the road users at the particular moment in the experiment. Therefore only the image frame from that camera is shown in (a). In (b), where the road users are tracked in 3D space, the CAV itself is labelled as an orange pillar with confidence ellipse and arrow indicating its moving direction. The two pedestrians are denoted by magenta pillars.}
	\label{fig:carla_irsu} 
\end{figure*}

\begin{figure*}[!t]
	\centering
	\subfloat[]{ 
		\label{fig:carla_cav_path:a} 
		\includegraphics[width=2.5in]{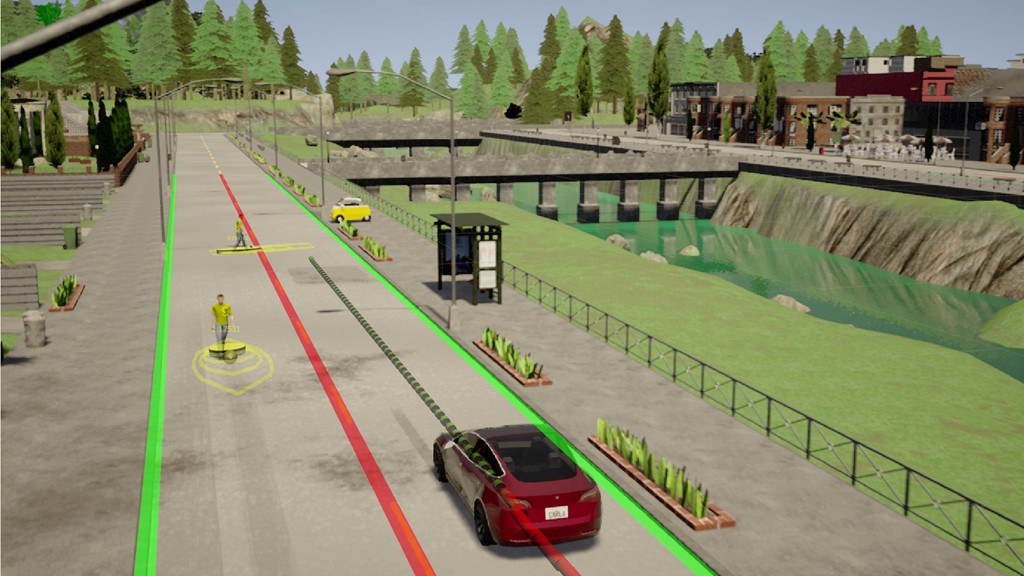}}
	\subfloat[]{ 
		\label{fig:carla_cav_path:b} 
		\includegraphics[width=2.5in]{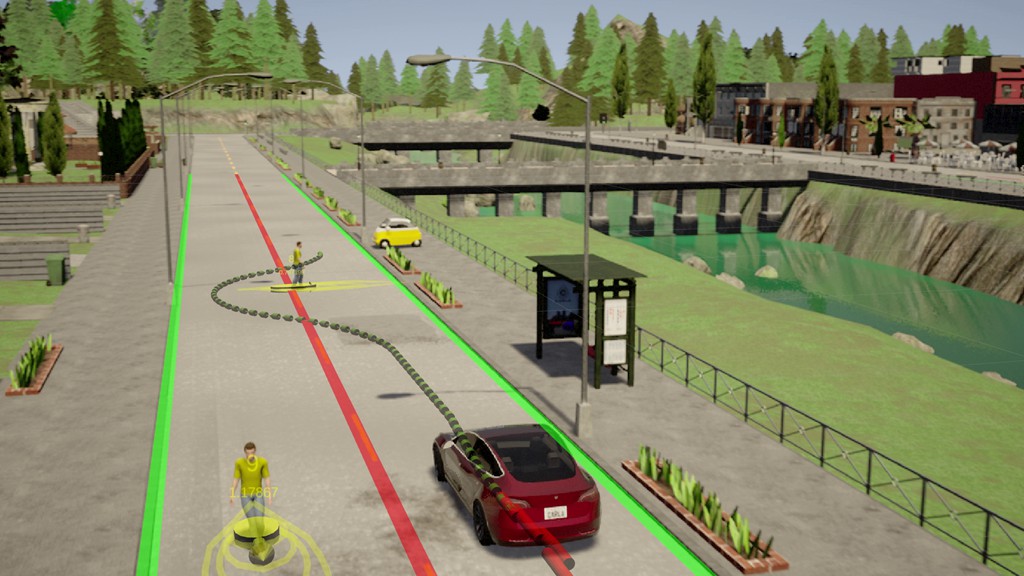}}
	\vfill
	\subfloat[]{ 
		\label{fig:carla_cav_path:c} 
		\includegraphics[width=5.1in]{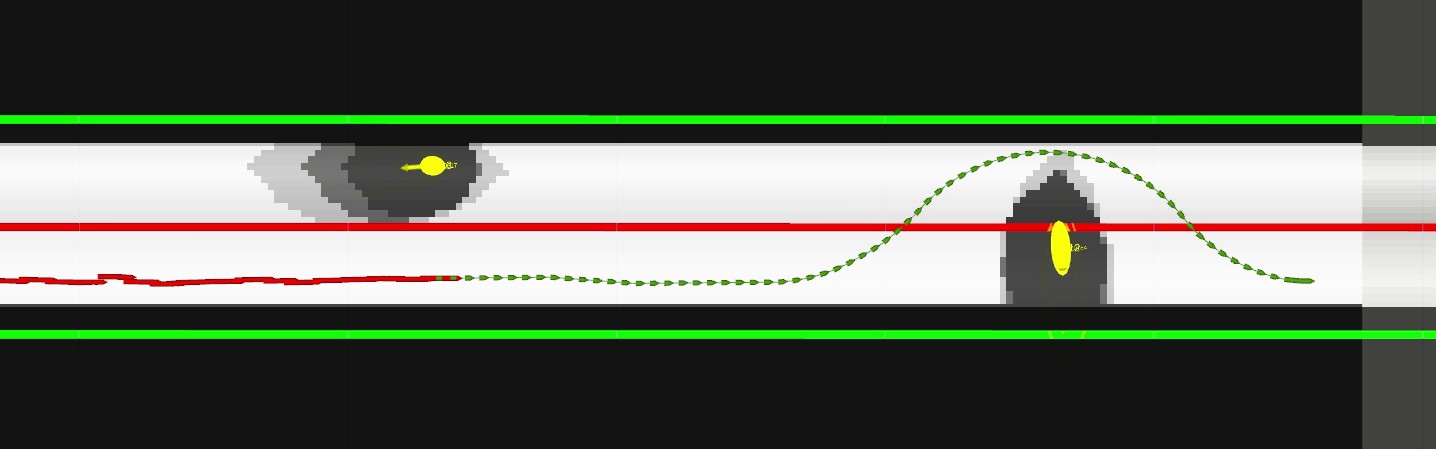}}
	\caption{The pedestrian tracking in the CAV relying only on CPMs from the IRSU in the CARLA simulator. In (c), the grid-based cost map maintained by the path planner in the CAV is presented with white as free space and black as occupied areas, and the odometry trajectory is shown as a red line, and the planned path is presented as a dotted green line.}
	\label{fig:carla_cav_path} 
\end{figure*}

In the experiment, as labelled in Figure \ref{fig:abercrombie_setup}, the CV first made a right turn from Meagher St to Abercrombie St, drove for about 60 m before turning to Little Queen St. As illustrated in Figure \ref{fig:abercrombie_irsu}, the CV learnt the ongoing traffic activity at the intersection with the IRSU deployed far beyond the reach of its onboard perception sensors. This illustrates the extended sensing range for smart vehicles through the CP service. At a later time shown in Figure \ref{fig:abercrombie_results:a} and Figure \ref{fig:abercrombie_results:b}, when the CV was about to make a right turn to Little Queen St, the vehicle could “see” a visually occluded pedestrian behind a building with the information coming through the IRSU. Please note that it is achieved seconds before the pedestrian could actually be visually picked up from the CV perspective in Figure \ref{fig:abercrombie_results:c} and Figure \ref{fig:abercrombie_results:d}, which is crucial for safety in both manual driving and autonomous driving scenarios. This is considered another benefit of the CP service demonstrated in the experiment.

Besides, the CV localisation quality varied depending on the lidar feature quality at different locations of the experiment environment. It is illustrated by comparing Figure \ref{fig:abercrombie_results:d} and Figure \ref{fig:abercrombie_results:f} that the refinement in egocentric pose estimate as a result of a correction in the self-localisation shown in the latter contributes to the improvement of tracking accuracy, which is consistent with the finding from the Section \ref{sec:num_sim} that the level of ITS-S localisation estimate uncertainty correlates to the perceived objects detection uncertainty after coordinate transformation and thus tracking uncertainty.

\subsection{Experiment in CARLA Simulator}

The experiment demonstrates the autonomous operation of a CAV using the perception information received from an IRSU in CARLA simulator as the only source for pedestrian estimation. CARLA \cite{paper:DosovitskiyRos2017} is an open source game engine based simulator, which supports flexible configurations of sensors, road users, and urban traffic environments. One of the highlights of CARLA simulator is the various and realistic sensory data it provides, including RGB images, lidar point clouds, IMU and GNSS measurements. Another highlight is its ROS integration with the \textit{ros-bridge} package. It is considered a safe, realistic, and repeatable environment for experimenting with autonomous driving systems, in particular, for conducting CAV operations that have safety concerns to test in the real world.

The experiment is setup with the IRSU detecting multiple walking pedestrians within its sensor range. The information of perceived pedestrians is then broadcast in the form of ETSI CPMs and is received by the CAV. The information is then transformed into its local frame of reference, and taken into account as the only perception data in the autonomous navigation of the CAV. The experiment is to demonstrate how a CAV interacts with VRU at non-designated crossing areas with the perception information from an IRSU. It is strongly recommended to conduct these experiments with higher risk CAV maneuvers in a simulator before performing them in real world.

The experiment was setup in the \textit{Town01} map of CARLA 0.9.8 simulator with participants labelled in Figure \ref{fig:carla_setup}. These include an IRSU deployed static next to a straight road, a CAV navigating autonomously on the same road towards east, and two walking pedestrians. One of the pedestrians was crossing the road, while the other is walking along the road in the west direction. A \textit{Lanelet2} map was built for the experiment environment in the simulator, which is essential for the CAV to stay on the road and plan drivable paths for autonomous navigation. The map includes the lane markings and the network of drivable regions as per right hand traffic rules. Accordingly, the CAV respects right hand traffic rules in the demonstration. On the road where the CAV is navigating, the CAV has right of way over pedestrians and is allowed to cross the broken dividing line to overtake other road users when it is safe to do so.

\begin{figure*}[!t]
	\centering
	\subfloat[]{ 
		\label{fig:carla_results:a} 
		\includegraphics[width=2.5in]{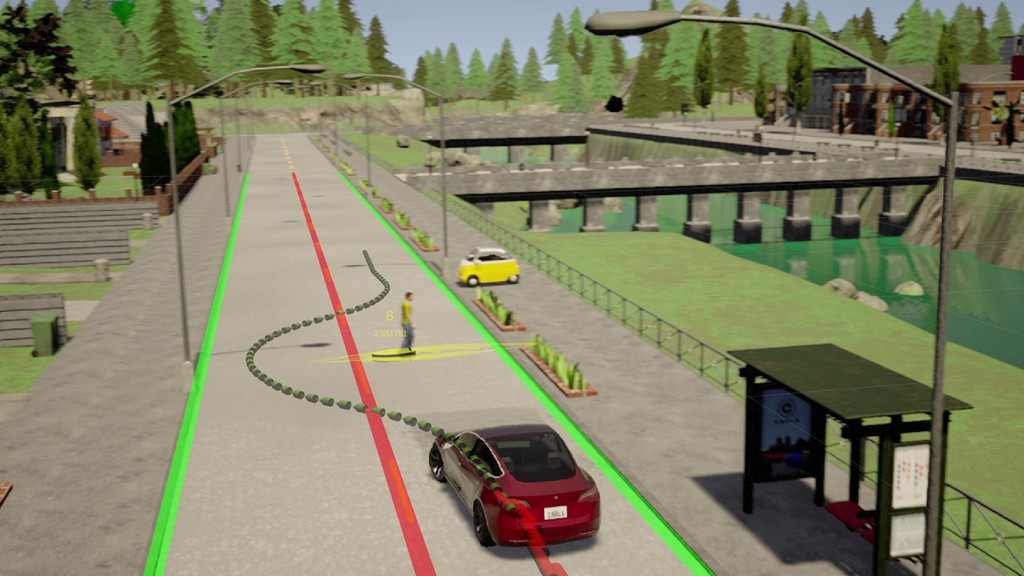}}
	\subfloat[]{ 
		\label{fig:carla_results:b} 
		\includegraphics[width=2.5in]{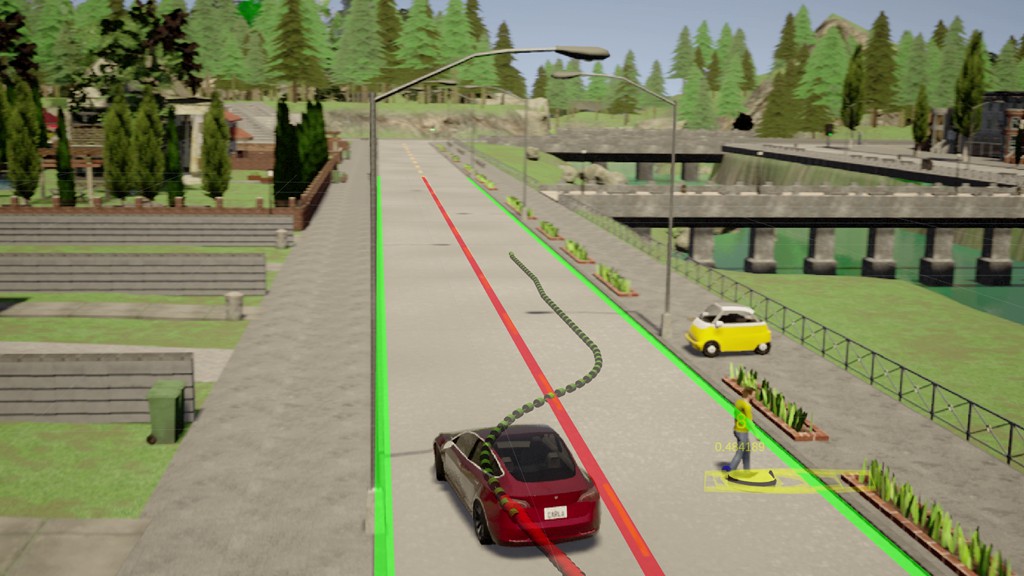}}
	\vfill
	\subfloat[]{ 
		\label{fig:carla_results:c} 
		\includegraphics[width=2.5in]{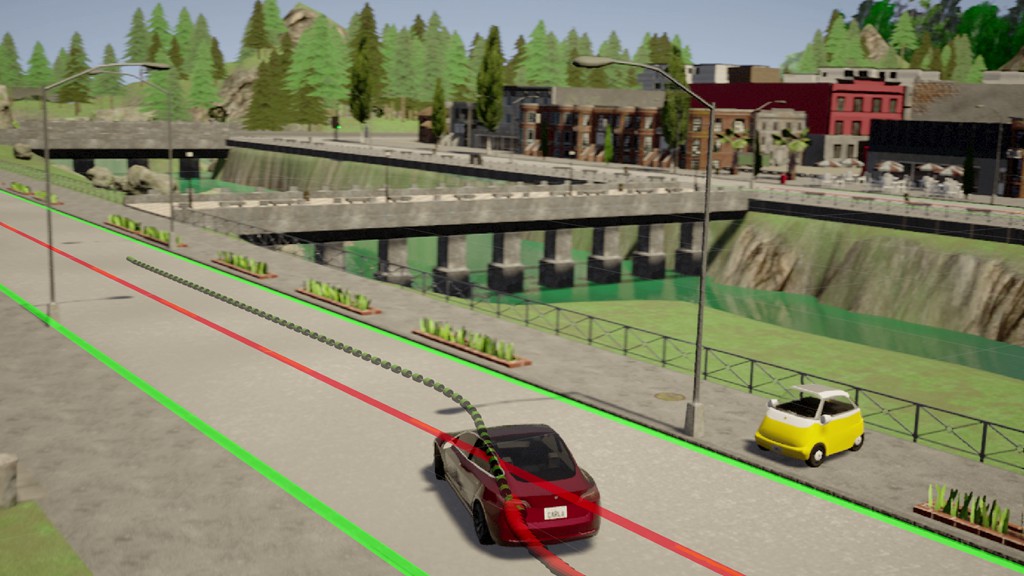}}
	\subfloat[]{ 
		\label{fig:carla_results:d} 
		\includegraphics[width=2.5in]{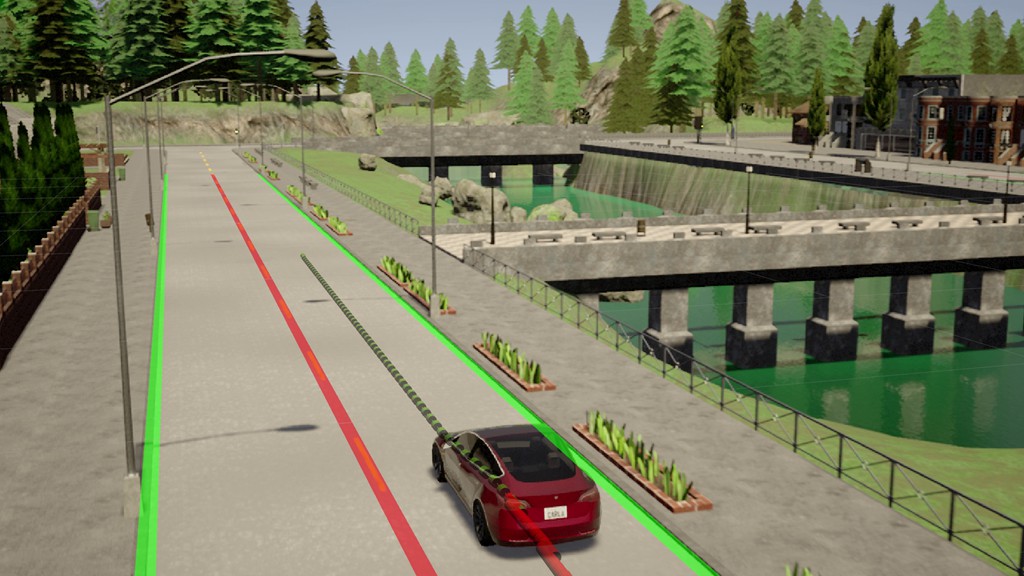}}
	\caption{The CAV safely navigated around the crossing pedestrian autonomously based on the CPMs from the IRSU in the CARLA simulator.}
	\label{fig:carla_results} 
\end{figure*}

The IRSU in the simulator is configured as close as possible to the real IRSU developed, which is equipped with two cameras and a 16-beam lidar, as revealed in Figure \ref{fig:carla_setup}. It is visualised as a mini car parked on the roadside in the simulator. The same suite of sensory data processing algorithms for the real IRSU is adopted in the simulated IRSU. The road user perception information is encoded to ETSI CPMs in the form of binary payloads and broadcast by the IRSU. The CAV employed in the experiment is as also shown in Figure \ref{fig:carla_setup}, which is able to position itself within the map using a combination of measurements from GNSS, IMU, and wheel encoder. The self-localization, however, is not assumed perfect but with noise added to the GNSS, IMU, and encoder measurements. 
A unscented Kalman filter (UKF) is used to achieve sub-meter accuracy.
In the experiment, the CAV is also setup to receive ETSI CPMs from the IRSU. It should be noted that the CAV does not have any onboard sensors for detecting road users, instead, it becomes aware of the road users and infer their states purely relying on information received from the IRSU. The CAV navigates at a speed up to 18 km/h in the simulator.

In the experiment, one of the pedestrians was crossing the road at a non-designated crossing area while the other is walking along the road in the opposite direction to the CAV. Figure \ref{fig:carla_irsu} demonstrates the road user detection and tracking in the IRSU at the beginning of the experiment. Specifically, Figure \ref{fig:carla_irsu:a} shows the detection of the two pedestrians and the CAV by fusing the lidar point cloud and image frame from Cam \#1 of the IRSU. The tracking of the same three road users in 3D space is shown in Figure \ref{fig:carla_irsu:b}. The tracking of the walking pedestrians in the CAV is illustrated in Figure \ref{fig:carla_cav_path}. Figure \ref{fig:carla_cav_path:c} depicts the planed path in the CAV to navigate around the pedestrians when approaching them. The tracked pedestrians are visualised with 95\% confidence ellipses and arrows indicating their moving direction. When the grid-based cost map is updated in the hybrid A* path planner in the CAV, the current and future estimates of the pedestrian states account for a higher cost (darker colour) in the grid such that the path planner would consider them when searching for an optimal and kinematically feasible path to its goal. Please note that the physical dimension of the CAV has been considered by bloating the confidence ellipses in the cost map.

In Figure \ref{fig:carla_cav_path}, it is shown that the CAV responded to the pedestrian walking along the road by not altering its planned path as expected. Although the pedestrian was walking on the road, it was blocking only the opposite lane of the road. In Figure \ref{fig:carla_cav_path:a}, the CAV became aware of the crossing pedestrian even when it was approximately 35 meters away, with the information coming through the IRSU. This was not taken into account by the path planning since the CAV  was still far from the pedestrian. As the CAV drove closer, it reacted to the event by planning a path that can safely navigate around the pedestrian while the pedestrian was crossing the road at a non-designated crossing area. As revealed in Figure \ref{fig:carla_results}, when the CAV was close to the pedestrian, it crossed the broken dividing lane to avoid the crossing pedestrian from behind. Following the avoidance maneuver, the CAV switched back to the right lane.

\subsection{Experiment in A Lab Environment}

\begin{figure*}[!t]
	\centering
	\includegraphics[width=4.0in]{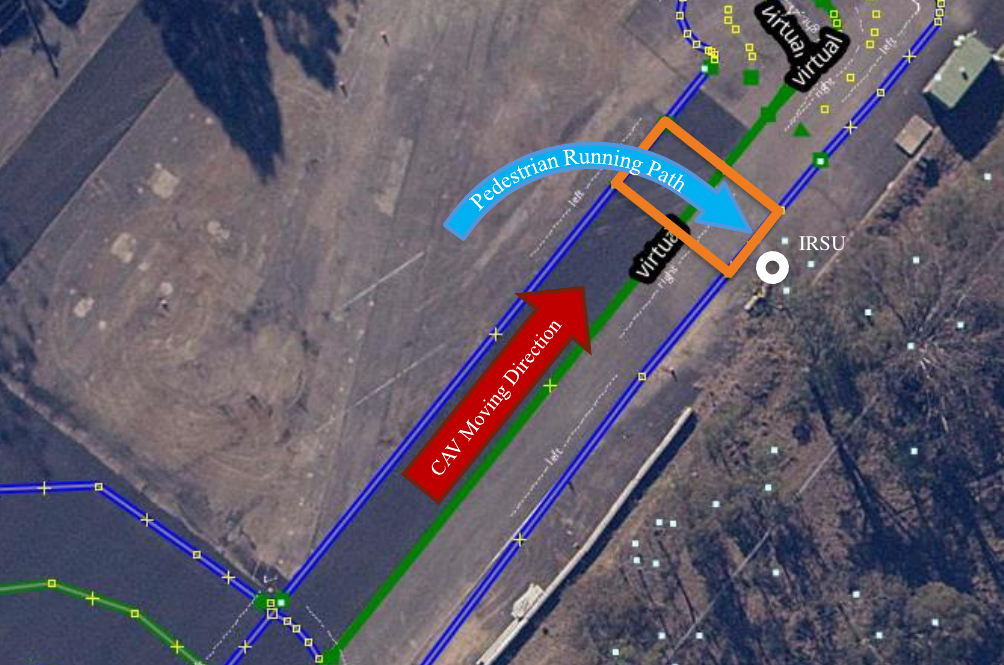}
	\caption{Experiment setup in a lab environment. It is shown in the \textit{Lanelet2} map and lidar feature map built for the experiment site, which contains the road boundaries and lanes information. The light blue dots distributed at the bottom right corner represent the locations of observed features in the environment (trees, poles, corners of structures). The designated pedestrian crossing area is labelled as an orange rectangle.}
	\label{fig:lab_setup}
\end{figure*}

The last experiment presented in the paper investigates the CAV interacting with a pedestrian running towards a pedestrian crossing area. It was setup in a lab traffic environment with a 55m stretch of straight road, a virtual pedestrian crossing area, and the IRSU deployed next to the crossing, as illustrated in Figure \ref{fig:lab_setup}. A \textit{Lanelet2} map was built for the environment and the CAV was able to localise itself within the pre-built map of lidar features, such as poles and building corners. The pedestrian crossing area was encoded in the \textit{Lanelet2} map which is stored locally in the CAV. The expected behaviour of the CAV is to drive through the crossing or stop before the crossing to give way depending on whether there is any pedestrian crossing activity detected by the IRSU.

\begin{figure*}[!t]
	\centering
	\subfloat[]{ 
		\label{fig:lab_irsu:a} 
		\includegraphics[width=2.4in]{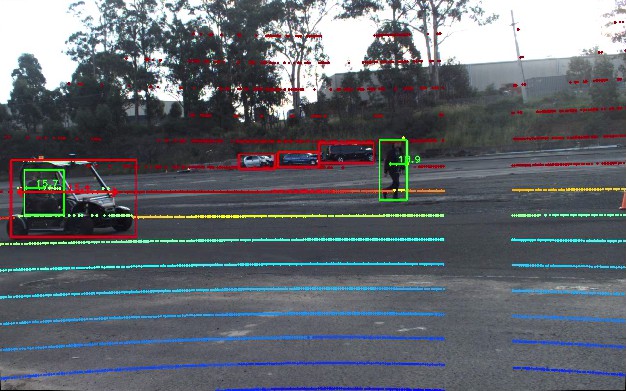}}
	\subfloat[]{ 
		\label{fig:lab_irsu:b} 
		\includegraphics[width=2.4in]{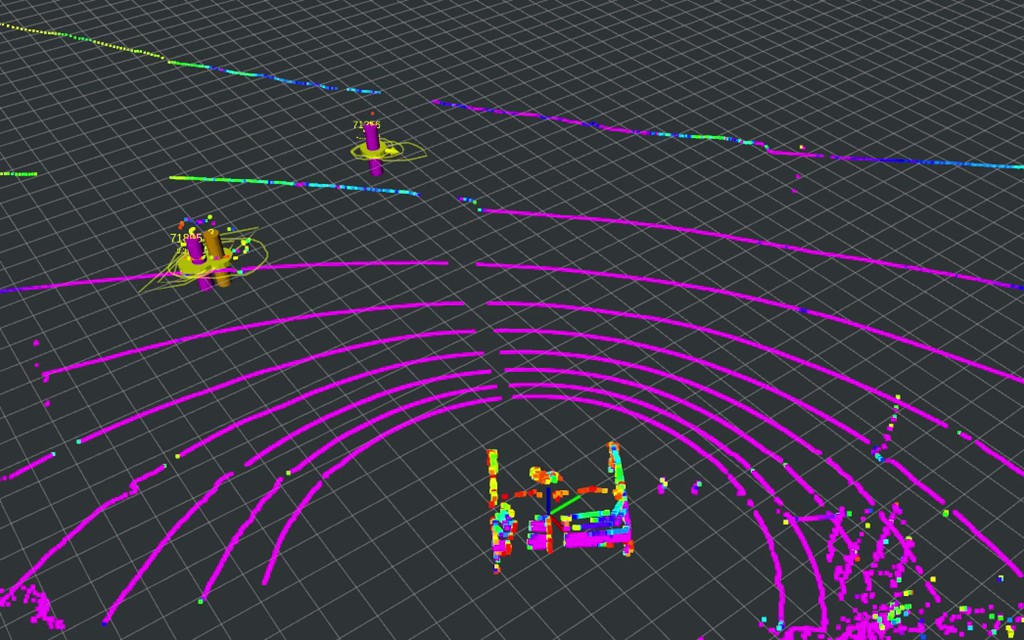}}
	\vfill
	\subfloat[]{ 
		\label{fig:lab_irsu:c} 
		\includegraphics[width=2.4in]{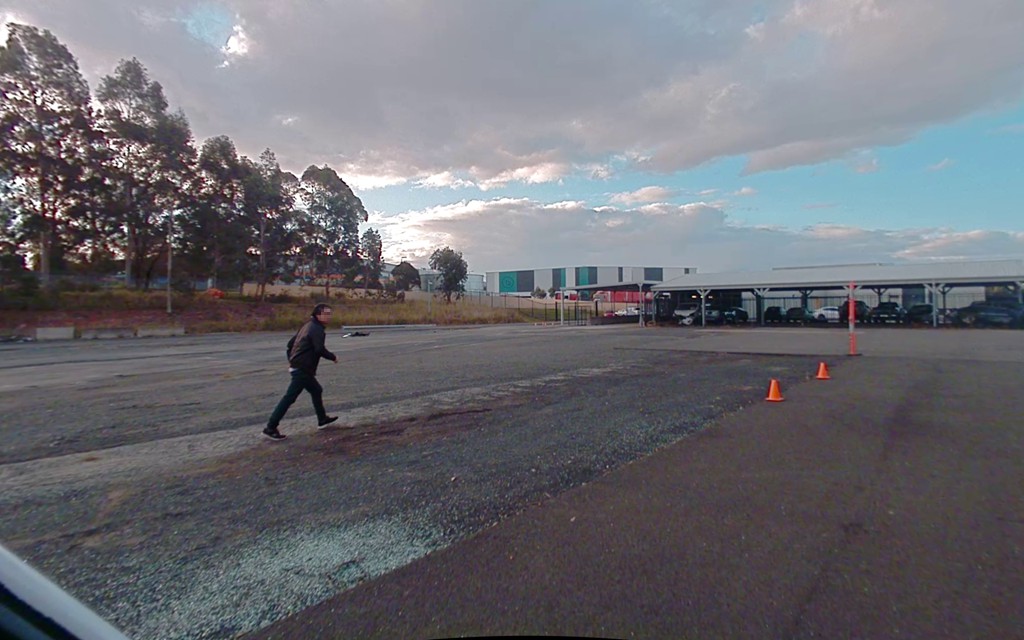}}
	\subfloat[]{ 
		\label{fig:lab_irsu:d} 
		\includegraphics[width=2.4in]{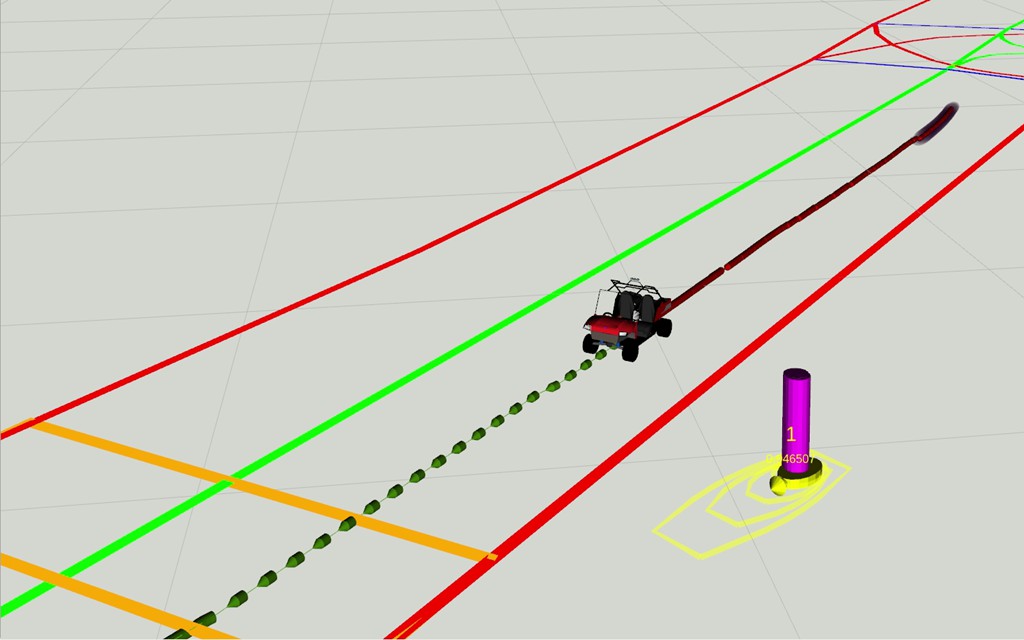}}
	\caption{The CAV started tracking a pedestrian running toward a crossing line of a marked pedestrian crossing in a lab environment based on ETSI CPMs from an IRSU. (a) shows the image frame from Cam \#1 of the IRSU with projected lidar point cloud. (b) and (d) illustrate the tracking of the road users in 3D space in the IRSU and the CAV, respectively, and (c) shows the image captured by the front-left camera on the CAV. The pedestrian and the safety driver of the CAV are tracked and denoted by magenta pillars with confidence ellipse and arrows indicating their moving direction, while the vehicle is labelled as an orange pillar in the IRSU's tracker. Only the running pedestrian was tracked in the GMPHD tracker running within the CAV, as depicted in (d), as the CAV itself and its safety driver are identified and removed from tracking based on the location of the received perceived objects after transformed into its local frame. The \textit{Lanelet2} map is also visualised in (d), where the pedestrian crossing is denoted by an orange rectangle. Furthermore, the odometry trajectory of the CAV is shown as a red line, and its planned path is denoted as a dotted green line in (d).}
	\label{fig:lab_irsu} 
\end{figure*}

\begin{figure*}[!t]
	\centering
	\subfloat[]{ 
		\label{fig:lab_results:a} 
		\includegraphics[width=2.4in]{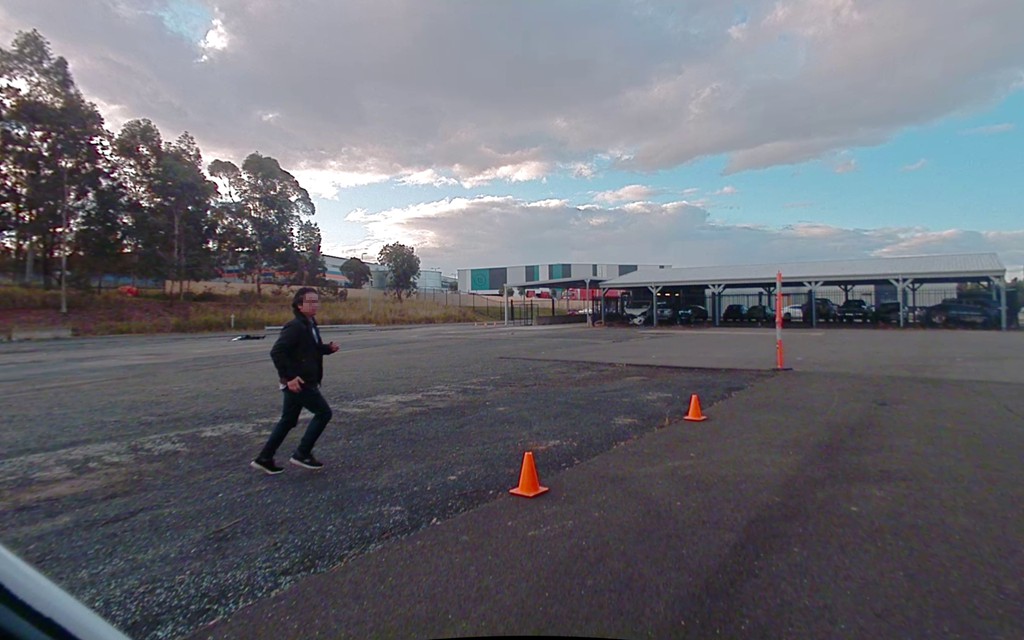}}
	\subfloat[]{ 
		\label{fig:lab_results:b} 
		\includegraphics[width=2.4in]{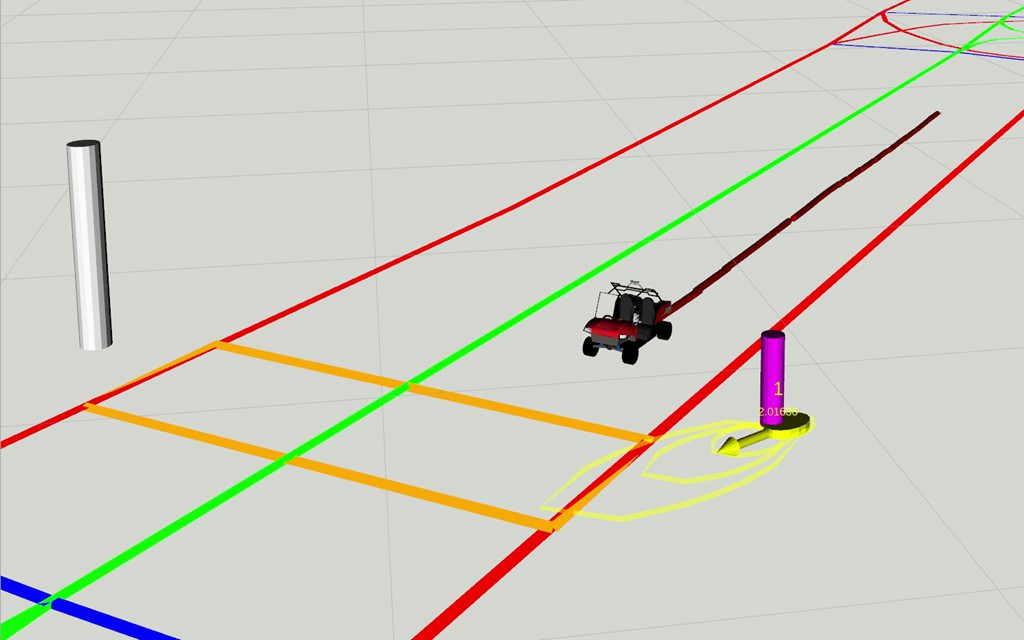}}
	\vfill
	\subfloat[]{ 
		\label{fig:lab_results:c} 
		\includegraphics[width=2.4in]{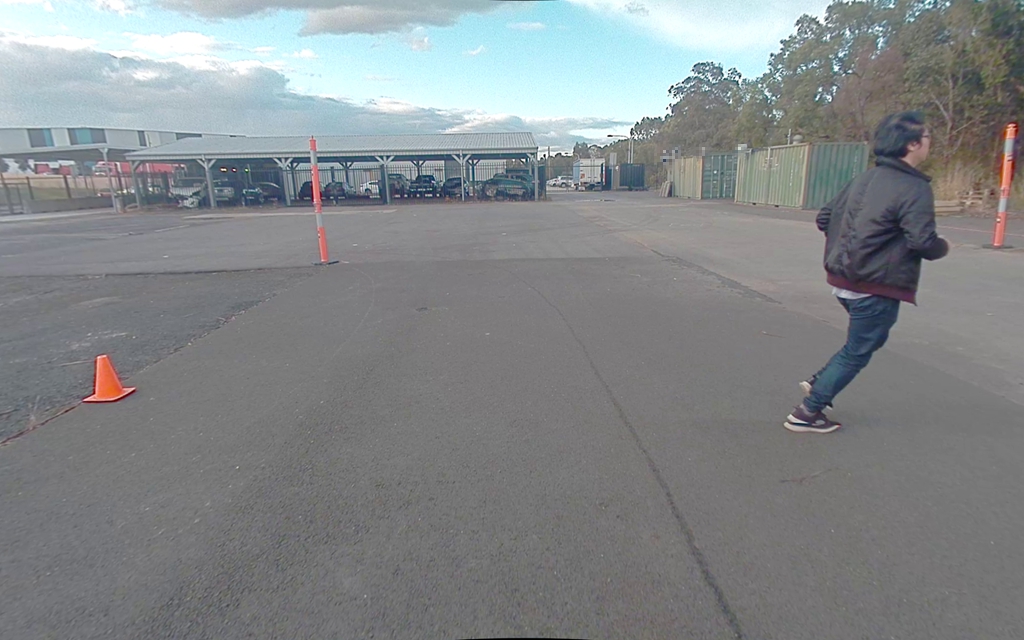}}
	\subfloat[]{ 
		\label{fig:lab_results:d} 
		\includegraphics[width=2.4in]{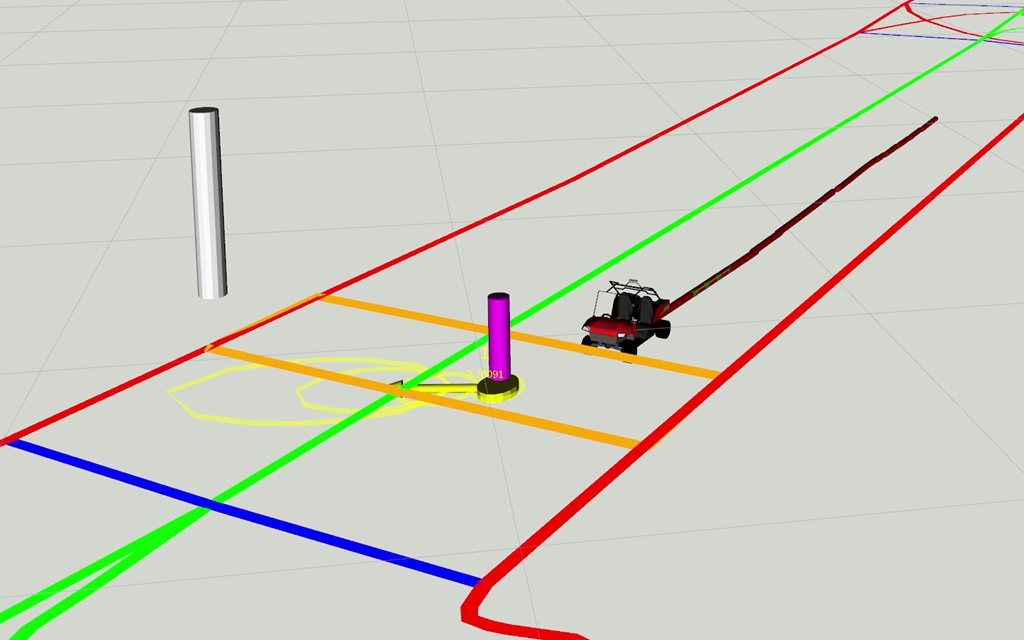}}
	\vfill
	\subfloat[]{ 
		\label{fig:lab_results:e} 
		\includegraphics[width=2.4in]{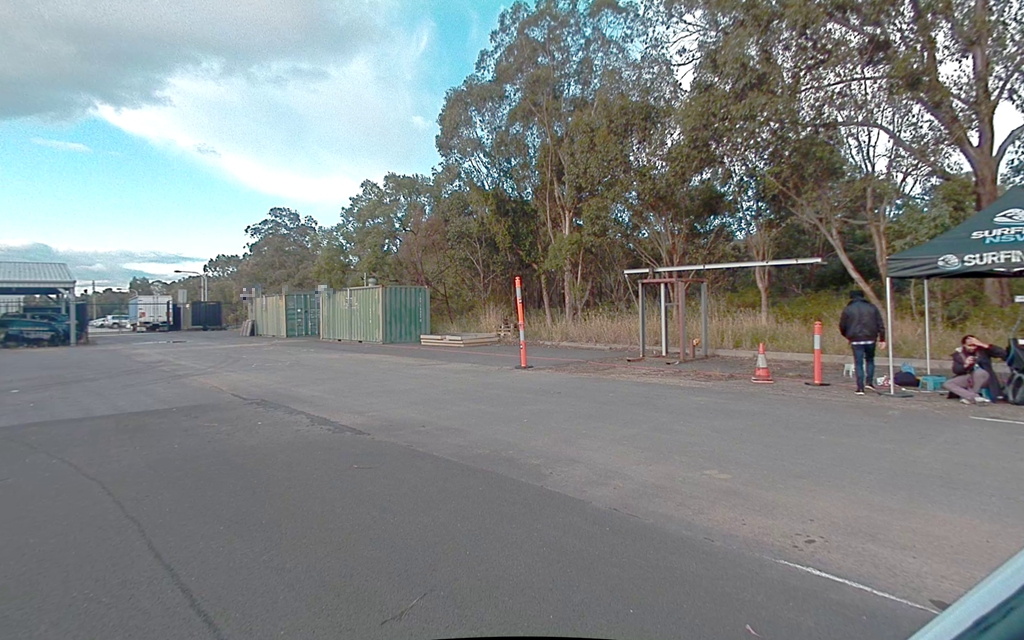}}
	\subfloat[]{ 
		\label{fig:lab_results:f} 
		\includegraphics[width=2.4in]{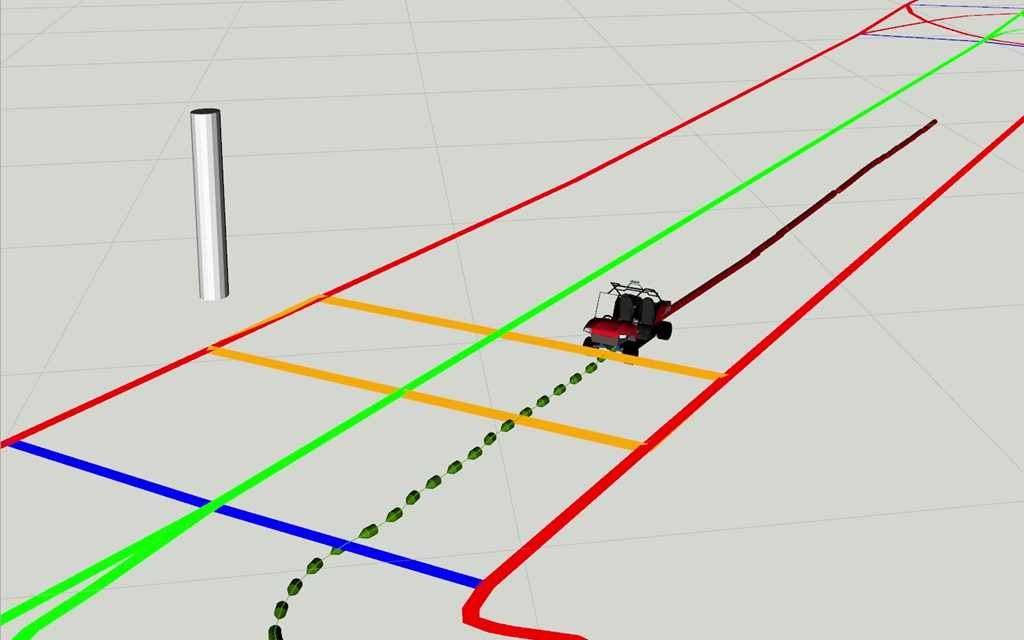}}
	\caption{The CAV gave way to a running pedestrian at the designated crossing area in a lab environment. Based on the ETSI CPMs broadcast by the IRSU, the CAV predicted the state of the running pedestrian up to 1.5 seconds into the future. The IRSU is visualised as a tall white pillar in the CAV in (b), (d), and (f). It is seen in (b) that the predicted states of the pedestrian, visualised as yellow ellipses, intersected with the orange crossing area on the \textit{Lanelet2} map. The CAV stopped for the pedestrian in front of the crossing in (d). The CAV then planned a new path forward after the pedestrian finished crossing in (f). In (a), (c), and (e), image frames from front-left, front-centre, front-right cameras on the CAV are presented, respectively. }
	\label{fig:lab_results} 
\end{figure*}

Figure \ref{fig:lab_irsu} shows the moment the pedestrian started running at the beginning of the experiment. The IRSU captured the road users in the scene, which include pedestrian and the CAV in 3D space. The CAV then received from the IRSU the information about the presence of a pedestrian running towards the crossing, as depicted in Figure \ref{fig:lab_irsu:d}. When approaching him, the CAV kept tracking and predicting the pedestrian state based on the perception information from the IRSU. A constant velocity kinematic model was adopted for predicting the future state of the pedestrian. As shown in Figure \ref{fig:lab_results:b}, the pedestrian was predicted by the CAV to occupy the crossing area, although he had not physically stepped onto the crossing yet. In the meantime, the CAV started to brake preemptively for the predicted crossing activity. The crossing area became a high cost region in the cost map within the path planner when the pedestrian was crossing. As a result, there was no feasible path found to drive through the crossing area for the CAV. It eventually stopped before the crossing line for the pedestrian, waiting until he finished crossing, as demonstrated in Figure \ref{fig:lab_results:d}. When the pedestrian crossing area was clear a moment later, the CAV re-planed a path to the original goal and started to drive through the crossing at a slightly slower speed due to the road speed rule encoded in the map, as depicted in Figure \ref{fig:lab_results:f}.

The CAV onboard perception sensors were not used for pedestrian detection throughout the experiment. Yet, there are a few measures in place to ensure safety when conducting autonomous navigation related activities in real world experiments. First of all, the experiment was conducted in a closed, controlled environment and the CAV speed is limited to 11 km/h when in autonomous mode. Besides, there is always a safety driver sitting in the CAV closely monitoring any hazards and has the ability to trigger an e-stop of the CAV in case of an emergency. Furthermore, the virtual bumper subsystem based on lidar points obstacle detection is enabled on the CAV, which brings the vehicle to an e-stop if any obstacle breaches any of the pre-set distance thresholds to the vehicle. However, the virtual bumper treats every surrounding object as static obstacles and therefore did not take part in the decision making process of the CAV in Figure \ref{fig:lab_results:b}, where it decided to give way based on the predicted future state of the pedestrian.

\section{Conclusions}
\label{sec:conclusions}

The paper is mainly focused on safety and robustness implications the CP can bring to the operations of CAVs. It presents the IRSU and CAV platforms developed by the joint research team from the ACFR and Cohda Wireless over the last two years, and demonstrates through three representative experiments the benefits associated with the use of CP service for CV and CAV operations alongside VRU in different traffic contexts.

In the first experiment conducted on a public road, the CV was able to track a pedestrian visually obstructed by a building with CP information from an IRSU deployed on the road side. This was achieved seconds before its local perception sensors or the driver could possibly see the same pedestrian around the corner, providing extra time for the driver or the navigation stack to react to a safety hazard. The second experiment conducted in CARLA simulator shows the autonomy of a CAV using the CP service. The experiment highlights the autonomous navigation of the CAV and its safe interaction with walking pedestrians, purely responding on the perception information provided by the IRSU. The last experiment demonstrated the expected behaviour of the CAV when interacting with a pedestrian rushing towards the designated crossing area in a lab environment. The CAV managed to take preemptive action, that is, braking and stopping before the crossing area for the pedestrian based on the kinematic prediction of the pedestrian. The pedestrian tracking, prediction, path planning and decision making in the CAV in the last experiment were operating based on the perception information received from the IRSU.

We believe these demonstrations can assist engineers and researchers in the relevant fields in achieving a better understanding of the safety implications the CP is bringing to the current and future transportation systems. The CP enables the smart vehicles to break the physical and practical limitations of onboard perception sensors, and in the meantime, to embrace improved perception quality and robustness along with other expected benefits from the CP service and V2X communication. As importantly, the CP can also reduce the reliance on the vehicle's local perception information, thereby lowering the requirement and cost for onboard sensing systems. Furthermore, it is demonstrated that when properly used, IRSU perception data can be used as another reliable source of information to add additional robustness and integrity to autonomous operations. These potentially trigger more in-depth discussions in the research community and industry related to the future development directions of intelligent vehicles and other C-ITS services and applications, such as maneuver intention sharing and cooperative driving.

\section{Acknowledgment}
The authors would like to thank HxGN SmartNet for providing Corrections and Data Service in the research work.

\bibliography{references}
\bibliographystyle{IEEEtran}

\end{document}